%% file: main.tex
\documentclass{article} 
\usepackage{iclr2027_conference,times}

\input{math_commands.tex}

\usepackage{hyperref}
\usepackage{url}
\usepackage{graphicx}
\usepackage{subcaption}
\usepackage{multirow}
\usepackage{booktabs}
\usepackage{array}
\usepackage[table]{xcolor}
\usepackage[most]{tcolorbox}
\usepackage{booktabs}

\definecolor{depthblue}{RGB}{29,151,249}

\title{Pumpire: Unified Benchmark for Metric Distance Estimation}

\author{
\textbf{Siyu Chen}$^{1, *}$ \quad
\textbf{Zehan Wang}$^{1, *, \dagger}$ \quad
\textbf{Jiayang Xu}$^{1, *}$ \quad
\textbf{Yihan Wu}$^{2}$ \quad
\textbf{Jialei Wang}$^{1}$ \\
\textbf{Junming Chen}$^{1}$ \quad
\textbf{Ziang Zhang}$^{1}$ \quad
\textbf{Yutong Ying}$^{3}$ \quad
\textbf{Zhou Zhao}$^{1,\ddagger}$ \\
$^{1}$Zhejiang University; $^{2}$Zhejiang University of Technology; $^{3}$Shanghai Jiaotong University 
}

\newcommand{\subfigref}[2]{\hyperref[#1]{\ref*{#1}#2}}

\newcounter{finding}
\renewcommand{\thefinding}{\arabic{finding}}   

\newenvironment{finding}{%
\refstepcounter{finding}%
\begin{tcolorbox}[
    enhanced,
    colback=gray!6!white,       
    colframe=gray!55!black,     
    boxrule=0.5pt,
    arc=2mm,
    auto outer arc,
    fuzzy shadow={
      1.5pt}{-1.5pt}{0pt}{0.5pt}{
      gray!100,opacity=0.35
    },
    left=8pt,
    right=8pt,
    top=1.5pt,
    bottom=1.5pt,
    before skip=5pt,
    after skip=5pt
  ]
\textbf{\textit{Finding \thefinding:}}\
}{%
\end{tcolorbox}
}

\newcommand{\compactspacing}{%
  \setlength{\tabcolsep}{2.5pt}%
  \renewcommand{\arraystretch}{1.05}%
}

\iclrfinalcopy 
\begin{document}

\maketitle
\lhead{}

\begingroup
\renewcommand{\thefootnote}{}
\footnotetext{%
  $^{*}$Equal contribution.\quad
  $^{\dagger}$ Project leader. \quad
  $^{\ddagger}$Corresponding author.
}
\endgroup

\input{sections/00_abstract}
\input{sections/01_introduction}
\input{sections/02_related_work}
\input{sections/03_data_pipeline}
\input{sections/04_protocol}
\input{sections/05_experiment}
\input{sections/06_insights}
\input{sections/07_conclusion}

\subsection*{AI use statement}

In this work, we used ChatGPT for programming assistance, language polishing, and visual refinement of paper figures. We did not use generative AI tools to generate research ideas, determine the technical methodology, collect the dataset, or annotate the dataset, and other tasks requiring disclosure are not applicable to this work. All AI-assisted text, code, and visual materials were reviewed and verified by the authors. We take responsibility for the final content of this work, including text, claims, code, figures, and other artifacts produced with the aid of generative AI.

\bibliography{main}
\bibliographystyle{iclr2027_conference}
\clearpage
\input{sections/10_appendix}

\end{document}

%% file: math_commands.tex
\usepackage{amsmath,amsfonts,bm}

\def\eqref#1{equation~\ref{#1}}

\def\1{\bm{1}}

\DeclareMathAlphabet{\mathsfit}{\encodingdefault}{\sfdefault}{m}{sl}
\SetMathAlphabet{\mathsfit}{bold}{\encodingdefault}{\sfdefault}{bx}{n}



%% file: sections/00_abstract.tex
\begin{abstract}
We present \textbf{\textit{Pumpire}}, a unified benchmark for evaluating metric point-pair distance estimation capability of both image- and video-level 3D foundation models, with or without depth priors. In contrast to previous approaches that normally evaluate depth and camera intrinsics separately or evaluate point-clouds with geometric similarity metrics, which cannot directly reflect models' point-to-point distance estimation capability, Pumpire directly assesses point-to-point distances from the reconstructed geometry. To this end, we collect a large-scale and diverse dataset (\textbf{\textit{pumpire-6k}}) comprising 100 real-world scenes, each annotated with physically measured point-pair distances and containing 64 frames, for a total of 6,400 frames. Building on this dataset, we establish a holistic evaluation protocol that covers both image- and video-level 3D foundation models and enables direct assessment of point-pair distance errors and cross-setting comparison. We conduct extensive experiments across 29 baseline configurations of representative 3D foundation models and provide a comprehensive analysis of the results. By offering this benchmark, we target the more fundamental ability to perceive and estimate physical scale in the reconstructed 3D space, which prior evaluation protocols have largely overlooked. The project page can be found at \url{https://pumpire.github.io/}
\end{abstract}

%% file: sections/01_introduction.tex
\section{Introduction} \label{sec:introduction}

Accurate estimation of metric distances between 3D point pairs—such as the physical separation between objects or individual object scale—has been a foundational capability across modern intelligent systems. In autonomous driving, translation errors between 3D detections and ground truth are associated with collision rates~\cite{taamazyan2024collision, caesar2020nuscenes}. In robotic manipulation, grasp success rates depend on 3D spatial measurements, not pixel-level depth accuracy alone~\cite{hietanen2021benchmarking, hu2026reliability}. In vision‑language models~\cite{cheng2024spatialrgpt, chen2024spatialvlm, shi2026realm, ma2025spatialllm, yang2025thinking}, limitations in accurately recovering metric distances and object scales remain a critical bottleneck for spatial reasoning and physical task execution. Hence, accurate metric point‑pair distance estimation serves as a core enabler of safe, reliable, and generalizable behavior.

Recent progress in metric 3D foundation models—exemplified by recent works~\cite{wang2507moge, hu2024metric3d, lin2511depth, leroy2024grounding, wang2026pi, ke2026depth}—has made it feasible to recover metric distances between arbitrary points in a scene from RGB inputs. These models can be categorized along two axes. The first concerns the input configuration: \textit{image-level} models operate on a single image, whereas \textit{video-level} models leverage multi-frame observations. The second concerns the prediction paradigm: \textit{estimation models} directly recover scene geometry from RGB inputs, while \textit{completion models} additionally condition on sparse depth priors to improve accuracy. Despite these advances, three critical gaps remain.

First, existing benchmarks for depth and camera intrinsics estimation are insufficient to characterize a model's point-pair distance estimation capability. Finding~\ref{find:tasks_coupling} shows that strong performance on depth and camera intrinsics benchmarks—even when considered jointly—does not necessarily translate into accurate metric distance estimation, highlighting the need for direct evaluation of point-pair distances.

Second, existing point-cloud benchmarks are also not designed to directly evaluate metric point-pair distance estimation. Current benchmarks primarily assess geometric fidelity using metrics such as Chamfer Distance, Accuracy, Completion, and F1-Score~\cite{wang2025vggt,leroy2024grounding,cabon2025must3r,wang2025continuous}, or point-wise reconstruction errors~\cite{wang2507moge}, rather than directly evaluating the accuracy of metric distance estimation between arbitrary point pairs. Meanwhile, recent spatial reasoning benchmarks~\cite{cheng2024spatialrgpt,chen2024spatialvlm,liao2024reasoning,guo2026surds,xu2026earthspatialbench} explicitly evaluate metric distance reasoning, but at a coarse, object-level granularity through question answering. Consequently, neither benchmark type directly assesses metric distance estimation at the point level, leaving the ability to recover physical distances between arbitrary point pairs insufficiently evaluated.

Third, there is no evaluation framework that enables unified comparisons of metric 3D foundation models across image- and video-level estimation and completion settings. Existing benchmarks are fragmented: some target image-level estimation~\cite{geiger2013vision,silberman2012indoor,vasiljevic2019diode,xian2024towards} or completion~\cite{wang2026aurora,li2025benchdepth,wong2020unsupervised,jung2023importance}, while others focus on video-level estimation~\cite{schroppel2022benchmark,cong2025e3d,schops2017multi,knapitsch2017tanks} or completion~\cite{dai2017scannet, shotton2013scene}. This fragmentation limits unified comparison of 3D foundation models across different settings.

In this work, we make the following contributions:
\begin{itemize}
    \item We introduce the first unified benchmark for directly evaluating the metric point-pair distance estimation capability of 3D foundation models, enabling unified comparisons across image- and video-level estimation and completion settings.
    
    \item We curate a real-world benchmark dataset comprising 100 diverse scenes and 6,400 frames, annotated with physically measured point-pair distances and spanning indoor and outdoor environments with varying levels of difficulty, enabling comprehensive evaluation of metric 3D foundation models under diverse real-world conditions.
    
    \item We conduct comprehensive evaluations across 29 baseline configurations of representative 3D foundation models and present empirical insights into metric point-pair distance estimation, revealing the strengths and limitations of different model categories and guiding downstream applications.
\end{itemize}

\begin{figure}[t]
    \centering
    \includegraphics[width=\linewidth]{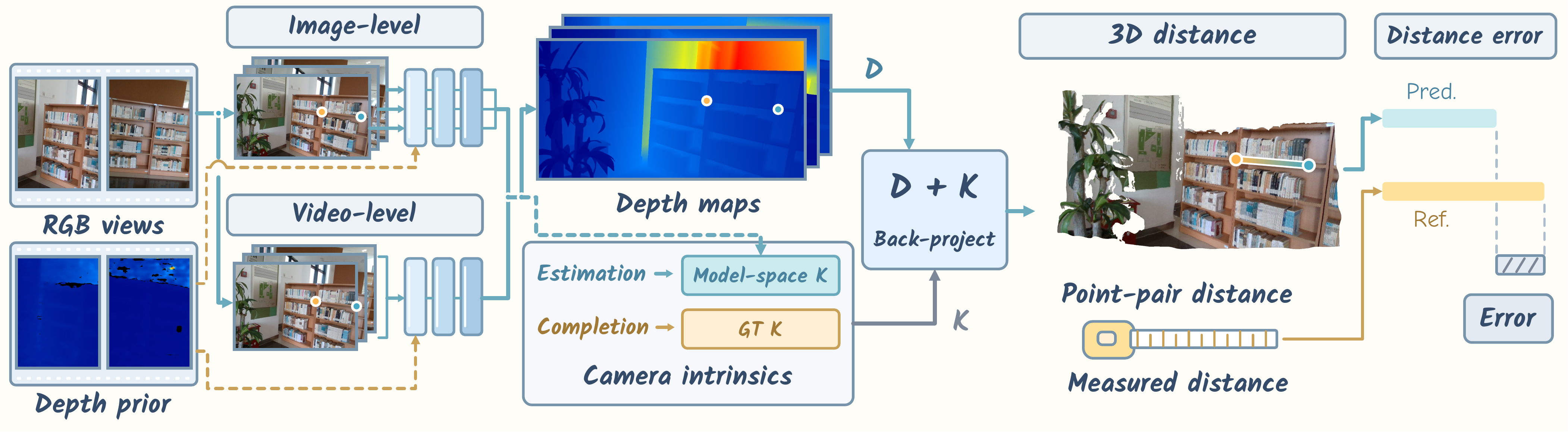}
    \caption{\textbf{Overview of the Pumpire evaluation protocol.} Pumpire provides a unified evaluation of image- and video-level 3D foundation models under geometry estimation and completion settings. Estimation models recover geometry from RGB inputs, whereas completion models additionally condition on depth priors. Pumpire back-projects depth maps predicted by 3D foundation models with their associated camera intrinsics and quantifies the results with the proposed metrics. Their predicted distance is compared with the physically measured ground truth using the proposed metrics as introduced in Section ~\ref{sec:benchmark_protocol}.}
    \label{fig:evaluation}
\end{figure}

%% file: sections/02_related_work.tex
\section{Related Work}
\label{sec:related_work}

\subsection{Geometry Estimation}

  Recent 3D foundation models have advanced both cross-domain generalization and metric-scale recovery from visual
  observations.
  At the image level, MiDaS~\cite{ranftl2020towards} and Depth Anything v2~\cite{yang2024depth} improve generalization
  through diverse training data, while Metric3D family~\cite{yin2023metric3d, hu2024metric3d} and
  Unidepth~\cite{piccinelli2024unidepth, piccinelli2025unidepthv2} explicitly account for camera geometry to recover
  metric depth.
  MoGe-2~\cite{wang2507moge} instead represents metric geometry through point maps.
  Beyond single-image prediction, video-level methods exploit geometric constraints across observations.
  DUSt3R~\cite{wang2024dust3r} reconstructs geometry from image pairs, and MASt3R~\cite{leroy2024grounding} augments this
  formulation with learned correspondences.
  Extending geometry prediction to multiple views, MUSt3R~\cite{cabon2025must3r}, VGGT~\cite{wang2025vggt}, and Depth Anything
  3~\cite{lin2511depth} infer consistent geometry across multiple observations, while CUT3R~\cite{wang2025continuous}
  adopts a stateful recurrent formulation that continuously updates a persistent scene representation.

  \subsection{Geometry Completion with Depth Priors}

  Geometry completion additionally incorporates depth measurements as metric constraints, requiring methods to handle
  incomplete observations and sensor noise.
  At the image level, CDM~\cite{liu2025manipulation} addresses camera-specific noise, while Lingbot-Depth~\cite{tan2026masked} refines incomplete and noisy depth through masked depth modeling.
  To accommodate varying depth coverage, Omni-dc~\cite{zuo2025omni} emphasizes generalization across sparsity patterns,
  and PriorDA~\cite{wang2025depth} combines incomplete metric measurements with dense relative depth predictions.
  At the video level, MapAnything~\cite{keetha2026mapanything} and pi3x~\cite{wang2026pi} support both RGB-only and depth-
  conditioned inference, bridging estimation and completion.
  Beyond direct input conditioning, CAPA~\cite{ke2026depth} incorporates depth observations through parameter-efficient
  test-time adaptation.
  These diverse input configurations and inference strategies motivate a unified evaluation of metric point-pair distance
  estimation, which Pumpire provides to inform model comparison for real-world applications.

%% file: sections/03_data_pipeline.tex
\section{pumpire-6K} \label{sec:pumpire_6k}

\begin{figure}[t]
    \centering
    \includegraphics[width=\linewidth]{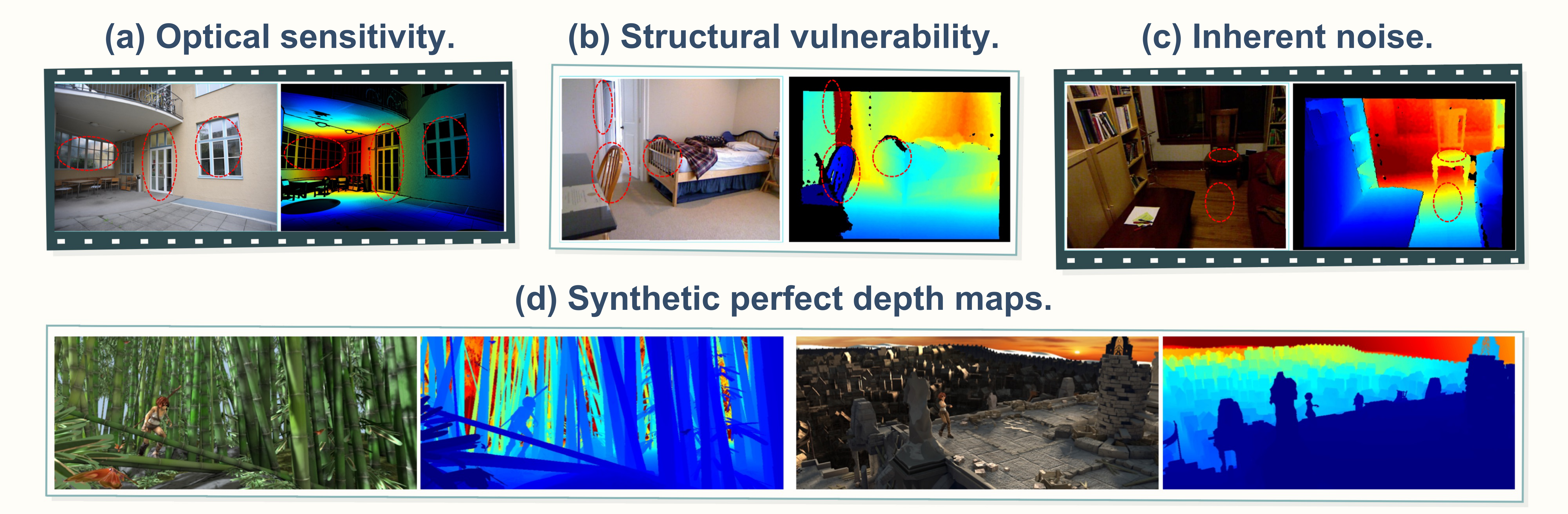}
    \caption{Limitations of existing depth benchmarks.}
    \label{fig:depth_limitations}
\end{figure}

\textbf{Limitations of current benchmarks.} In existing real-world datasets, point-pair distances derived from sensor-based depth maps can be inaccurate on reflective or transparent surfaces (Figure~\subfigref{fig:depth_limitations}{a}), thin structures and object boundaries (Figure~\subfigref{fig:depth_limitations}{b}), and even ordinary surfaces affected by inherent sensor noise (Figure~\subfigref{fig:depth_limitations}{c}). Synthetic datasets provide noise-free geometry but may not reflect real-world performance due to the domain gap (Figure~\subfigref{fig:depth_limitations}{d}). We therefore collect real-world scenes and manually measure physical point-pair distances with a tape measure.

\textbf{Large-scale collection.} We collect pumpire-6k from 100 diverse real-world scenes, including 40 indoor and 60 outdoor scenes. For each scene, we capture a sequence of 200--300 frames and uniformly sample 64 frames to reduce inter-frame redundancy, yielding 6,400 frames in total. Each sequence provides $1280\times720$ RGB images, RGB-aligned depth maps, camera intrinsics, the physically measured distance between two sticker centers, and their pixel coordinates in each frame. We annotate the sticker coordinates using a coarse-to-fine pipeline that combines manual anchor-frame annotation, CoTracker3-based propagation~\cite{karaev2025cotracker3}, and frame-by-frame verification. Further details are provided in Appendix~\ref{app:annotation_pipeline}.

\textbf{Diverse surface coverage.} As discussed above, sensor-derived depth annotations can be unreliable under challenging optical properties, texture patterns, and geometric structures. By physically measuring point-pair distances, we prevent these sensing errors from propagating into the ground truth, allowing pumpire-6k to cover a broad range of such conditions for comprehensive model evaluation. Table~\ref{tab:scene_materials} summarizes this coverage.

\begin{table}[h]
  \centering
  \caption{\textbf{Surface coverage of pumpire-6k.}
  ``N'' denotes the number of scenes containing each property. Categories are not mutually exclusive, as a scene may exhibit multiple properties.}
  \label{tab:scene_materials}
  \begin{tabular}{lclclc}
    \toprule
    Optical properties & N &
    Texture patterns & N &
    Geometric structures & N \\
    \midrule
    Common & 84 &
    Common & 83 &
    Common & 63 \\
    Transparent surfaces & 12 &
    Repetitive textures & 13 &
    Slender structures & 25 \\
    Reflective surfaces & 10 &
    Textureless regions & 5 &
    Near sharp edges & 11 \\
    \bottomrule
  \end{tabular}
\end{table}

\textbf{Randomized viewpoint acquisition.} During capture, we randomize the camera trajectories to ensure that the sequences encompass a wide variety of motion modes, exemplified by randomized distance scaling mode (bidirectional near–far transitions), variations in camera orientation, stochastic in-image positioning of stickers, and perspective variations across a wide range of viewing angles and distances. To adapt to the substantial depth variations that arise from such randomized motions, we employ circular stickers of 10 mm in radius, enabling reliable localization across the full range of capture distances.

Together, pumpire-6k combines manually measured point-pair distances and frame-verified pixel annotations with broad coverage of indoor and outdoor environments, local surface properties, and viewing conditions, enabling reliable and comprehensive evaluation of metric point-pair distance estimation.

%% file: sections/04_protocol.tex
\section{Pumpire Protocols}  \label{sec:benchmark_protocol}

As illustrated in Figure~\ref{fig:evaluation}, we introduce a unified protocol for evaluating image- and video-level metric 3D models under both estimation and completion settings. For models that directly predict a depth map, we use it as $D$; otherwise, we obtain $D$ from the $z$-coordinates of the predicted point-map. We then back-project the two annotated pixels into the camera coordinate system and compute their Euclidean distance:
\begin{equation}
d_{\mathrm{pred}}
= \left\|\mathbf{P}_0-\mathbf{P}_1\right\|_2,
\qquad
\text{where}\quad
\mathbf{P}_i
= D(u_i,v_i)K^{-1}[u_i,v_i,1]^\top,
\quad i\in\{0,1\}.
\end{equation}

Here, $K$ denotes the camera intrinsics, and the ground-truth distance $d_{\mathrm{gt}}$ is obtained through physical measurement as described in Section~\ref{sec:pumpire_6k}.

Let $S$ denote the number of scenes and $M_s$ the number of evaluated frames in scene $s$. Since the point pair remains fixed within each scene, $d_{\mathrm{gt}}^{(s)}$ is shared across all frames. We evaluate each scene using Absolute Error (AE), Relative Error (RE), Logarithmic Error (LE), and Threshold Accuracy ($\delta_\tau$): Define the frame-wise absolute error and distance ratio as $e_{s,j}=\left|d_{\mathrm{pred}}^{(s,j)}-d_{\mathrm{gt}}^{(s)}\right|$ and $r_{s,j}=d_{\mathrm{pred}}^{(s,j)}/d_{\mathrm{gt}}^{(s)}$. The scene-level metrics are then

  \noindent
  \begin{minipage}{0.48\linewidth}
  \begin{equation}
  \mathrm{AE}_s
  =\frac{1}{M_s}\sum_{j=1}^{M_s}e_{s,j}.
  \end{equation}
  \end{minipage}\hfill
  \begin{minipage}{0.48\linewidth}
  \begin{equation}
  \mathrm{RE}_s
  =\frac{1}{M_s}\sum_{j=1}^{M_s}
  \frac{e_{s,j}}{d_{\mathrm{gt}}^{(s)}}.
  \end{equation}
  \end{minipage}

  \noindent
  \begin{minipage}{0.48\linewidth}
  \begin{equation}
  \mathrm{LE}_s
  =\frac{1}{M_s}\sum_{j=1}^{M_s}
  \left|\log r_{s,j}\right|.
  \end{equation}
  \end{minipage}\hfill
  \begin{minipage}{0.48\linewidth}
  \begin{equation}
  \delta_{\tau,s}
  =\frac{1}{M_s}\sum_{j=1}^{M_s}
  \mathbf{1}\!\left[
  \max\!\left(r_{s,j},r_{s,j}^{-1}\right)<\tau
  \right].
  \end{equation}
  \end{minipage}

  The benchmark-level result for each metric
  $q\in\{\mathrm{AE},\mathrm{RE},\mathrm{LE},\delta_\tau\}$ is obtained by
  averaging its scene-level values:
  \begin{equation}
      q=\frac{1}{S}\sum_{s=1}^{S}q_s.
  \end{equation}

To evaluate cross-view consistency, we utilize Relative Standard Deviation
  (RSD) to measure the normalized variation of predicted distances within each
  sequence. For sequence $s$, RSD is defined as

  \begin{equation}
      \mathrm{RSD}_s
      =
      \frac{1}{\bar d_{\mathrm{pred}}^{(s)}}
      \sqrt{
      \frac{1}{M_s-1}
      \sum_{j=1}^{M_s}
      \left(
      d_{\mathrm{pred}}^{(s,j)}
      -\bar d_{\mathrm{pred}}^{(s)}
      \right)^2
      },
      \qquad
      \bar d_{\mathrm{pred}}^{(s)}
      =
      \frac{1}{M_s}
      \sum_{j=1}^{M_s}d_{\mathrm{pred}}^{(s,j)}.
  \end{equation}

%% file: sections/05_experiment.tex
\section{BENCHMARKING METRIC 3D FOUNDATION MODELS} \label{sec:systematic_evaluation}

\begin{table}[t]
\centering
\caption{\textbf{Performance of raw sensor depth on normal and challenging
  scenes.} $N$ denotes the number of scenes. AE is reported in meters, and
  $\delta$ metrics are reported as percentages.}
\label{tab:challenge_normal_compare}
\small
\begin{tabular}{l c c c c c c c c}
\toprule
  & N & AE & RE & LE & RSD & $\delta_{1.05}$ & $\delta_{1.10}$ & $\delta_{1.25}$ \\
\midrule
Normal    & 76  & 0.026 & 0.038 & 0.036 & 0.062 & 79.32 & 92.56 & 98.66 \\
Challenging & 24  & 4.047 & 8.390 & 1.228 & 0.887 & 21.03 & 27.60 & 36.72 \\
\midrule
Overall   & 100 & 0.991 & 2.043 & 0.322 & 0.260 & 65.33 & 76.97 & 83.80 \\
\bottomrule
\end{tabular}
\end{table}

\subsection{Scene Difficulty Split via Cross-View Consistency}
  \label{sec:data_split}

\begin{figure*}[t]
    \centering
    \includegraphics[width=\linewidth]{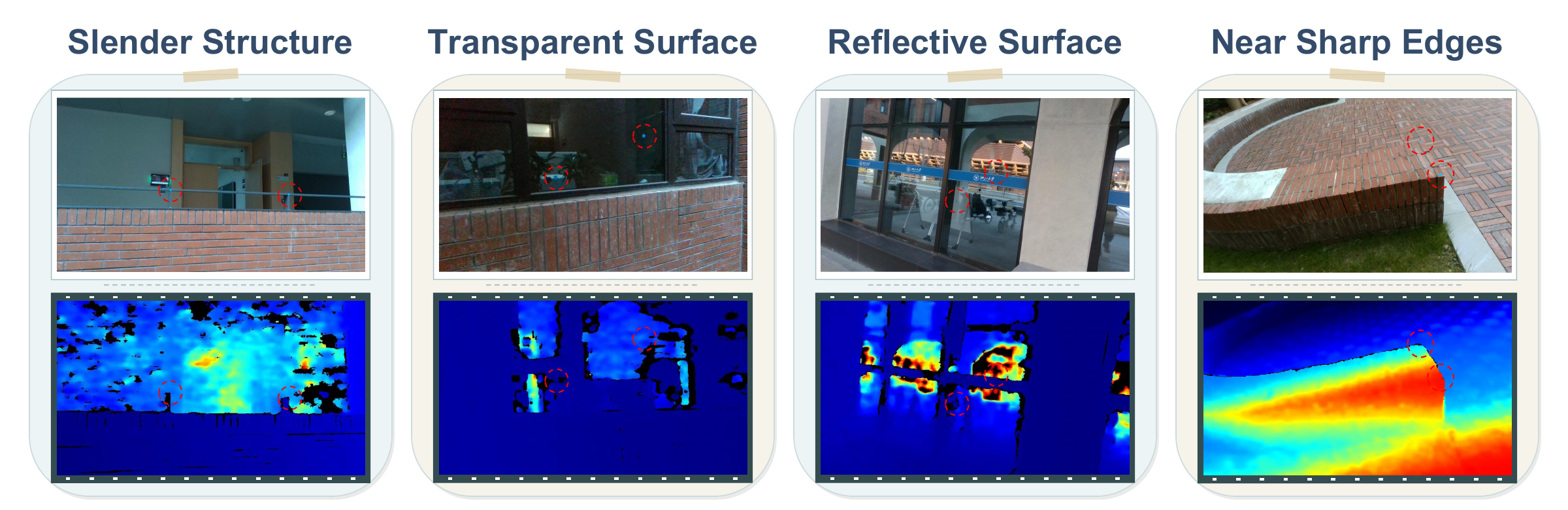}
    \caption{\textbf{Representative challenging scenes with unreliable sensor depth.} These cases correspond to the challenging subset identified via the RSD threshold ($> 0.1$).}
    \label{fig:challenge_sensor}
\end{figure*}

Severe failures on a small number of difficult scenes can dominate aggregate metrics and obscure performance on typical cases. We therefore report results separately on \emph{normal} and \emph{challenging} scenes. Since raw depth sensors are affected by optical and geometric conditions that also challenge learned 3D models, we use sensor reliability as a proxy for scene difficulty. Specifically, we measure the cross-view variation of point-pair distances derived from raw sensor depth using $\mathrm{RSD}_{\mathrm{sensor}}$, with higher values indicating less reliable depth measurements. Using an empirical threshold of $0.1$, we identify 24 challenging scenes and 76 normal scenes. Table~\ref{tab:challenge_normal_compare} summarizes the sensor performance on both subsets, and Figure~\ref{fig:challenge_sensor} shows representative challenging scenes.

\subsection{Quantitative results for 3D foundation models on our benchmark}

\textbf{Baselines.} We benchmark 29 representative baseline configurations spanning four settings: image-level estimation (8), image-level completion (10), video-level estimation (6), and video-level completion (5). All evaluated models predict scene geometry at absolute scale, enabling direct comparison under our unified distance-evaluation protocol. The complete model list, including the evaluated variants and backbones, is provided in Appendix~\ref{app:baselines}.

\textbf{Back-projection protocol.} As later demonstrated in Finding~\ref{find:tasks_coupling}, point-pair distance estimation depends on the coupling between depth and camera intrinsics. We therefore back-project predicted depth using the intrinsics associated with its camera space: predicted intrinsics (or canonical intrinsics for Metric3D~\cite{yin2023metric3d} and Metric3D v2~\cite{hu2024metric3d}) for estimation models, and ground-truth intrinsics for completion models conditioned on sensor depth. For video-level evaluation, we sample 32 groups of 10 frames per scene and average the metrics across all predictions. Video-level completion models additionally receive sensor-depth priors, with both depth and intrinsics provided to MapAnything~\cite{keetha2026mapanything} and pi3x~\cite{wang2026pi}.

\input{tables/main_mono}

\textbf{Image-level geometry estimation models.}
  As shown in Table~\ref{tab:mono}, Metric3D v2~\cite{hu2024metric3d} achieves the best average rank on the normal subset,
  followed by Unidepthv2~\cite{piccinelli2025unidepthv2} and MetricAnything~\cite{ma2026metricanything}.
  On the challenging subset, Depth Pro~\cite{bochkovskiy2025depth} and MoGe-3~\cite{kong2026moge} rank first and second,
  respectively, while Metric3D v2 drops substantially in average rank.
  These results suggest that strong performance under normal conditions does not necessarily generalize to challenging
  scenes.

\input{tables/main_mvs}

\textbf{Video-level geometry estimation models.}
  As shown in Table~\ref{tab:multi_view}, Depth Anything 3~\cite{lin2511depth} ranks first on both subsets.
  On the challenging subset, pi3x~\cite{wang2026pi} improves from fourth to second in average rank, whereas
  MUSt3R~\cite{cabon2025must3r} drops from third to fifth. Both pi3x and MapAnything~\cite{keetha2026mapanything} support depth-prior conditioning, with the corresponding completion results reported in Table~\ref{tab:mvd_completion}.

\input{tables/main_mono_cmpl}

\textbf{Image-level geometry completion models.}
  As shown in Table~\ref{tab:completion}, Lingbot-Depth~\cite{tan2026masked} ranks first on both the normal and challenging
  subsets.
  The relative ordering of the other methods changes markedly across the two subsets.
  In particular, CDM-L515 and CDM-D435 ~\cite{liu2025manipulation} move from eighth and seventh place on the normal subset to
  second and third on the challenging subset, whereas Omni-dc~\cite{zuo2025omni} and LDCM~\cite{yu2026large}
  drop in average rank.

\input{tables/main_mvs_cmpl}

\textbf{Video-level geometry completion models.}
  As shown in Table~\ref{tab:mvd_completion}, the three CAPA variants~\cite{ke2026depth} achieve the top three average ranks
  on both subsets, with CAPA-VGGT ranking first.
  On the challenging subset, CAPA-MoGe2 achieves the lowest AE, RE, and RSD, whereas CAPA-VGGT achieves the lowest LE and
  the highest threshold accuracies.

%% file: tables/main_mono.tex
\begin{table}[t]
\centering
\caption{\textbf{Quantitative results for image-level geometry estimation.} The top-3 results are highlighted as \colorbox{green!30}{first}, \colorbox{yellow!30}{second}, and \colorbox{blue!20}{third}, the same highlighting applies below.}
\label{tab:mono}
\label{tab:mono_normal}
\label{tab:mono_challenge}
\compactspacing
\resizebox{\textwidth}{!}{%
\begin{tabular}{lcccccccc|cccccccc}
\toprule
& \multicolumn{8}{c|}{Normal}
& \multicolumn{8}{c}{Challenging} \\
\cmidrule(lr){2-9}\cmidrule(lr){10-17}
Model
& AE$\downarrow$ & RE$\downarrow$ & LE$\downarrow$ & RSD$\downarrow$
& $\delta_{1.05}\uparrow$ & $\delta_{1.10}\uparrow$
& $\delta_{1.25}\uparrow$ & Rank$\downarrow$
& AE$\downarrow$ & RE$\downarrow$ & LE$\downarrow$ & RSD$\downarrow$
& $\delta_{1.05}\uparrow$ & $\delta_{1.10}\uparrow$
& $\delta_{1.25}\uparrow$ & Rank$\downarrow$ \\
\midrule
Metric3D
& 1.180 & 1.695 & 0.828 & 0.347 & 8.02 & 14.68 & 27.82 & 8.00
& 5.004 & 9.273 & 1.299 & 0.672 & 7.68 & 14.13 & 23.11 & 8.00 \\
MoGe-3
& 0.208 & 0.292 & 0.236 & \cellcolor{yellow!30}0.084
& 18.89 & 35.49 & 59.89 & 6.00
& \cellcolor{green!30}0.158 & \cellcolor{green!30}0.275
& \cellcolor{green!30}0.228 & \cellcolor{green!30}0.112
& 12.63 & 26.82 & 58.20 & \cellcolor{yellow!30}3.00 \\
Depth Pro
& \cellcolor{blue!20}0.135 & 0.196 & 0.234 & 0.154
& 17.74 & 32.98 & 65.30 & 5.57
& \cellcolor{yellow!30}0.171 & \cellcolor{blue!20}0.426
& \cellcolor{yellow!30}0.250 & \cellcolor{blue!20}0.220
& 15.82 & 31.58 & \cellcolor{green!30}68.10
& \cellcolor{green!30}2.86 \\
Unidepth
& 0.159 & 0.220 & 0.207 & 0.134
& 20.00 & 38.38 & \cellcolor{blue!20}68.44 & 4.86
& 0.943 & 1.941 & 0.676 & 0.337 & 15.10 & 26.24 & 42.90 & 6.29 \\
MoGe-2
& 0.176 & 0.242 & 0.204 & \cellcolor{blue!20}0.086
& 23.40 & 40.93 & 63.08 & 4.71
& \cellcolor{blue!20}0.202 & \cellcolor{yellow!30}0.355
& \cellcolor{blue!20}0.258 & \cellcolor{yellow!30}0.206
& 15.95 & 30.21 & 57.03 & \cellcolor{blue!20}3.43 \\
MetricAnything
& 0.145 & \cellcolor{blue!20}0.194 & \cellcolor{blue!20}0.176
& 0.089 & \cellcolor{blue!20}25.00
& \cellcolor{blue!20}41.28 & 66.96 & \cellcolor{blue!20}3.43
& 0.333 & 0.743
& 0.315 & 0.242
& \cellcolor{blue!20}18.68 & \cellcolor{blue!20}35.35
& \cellcolor{blue!20}64.84 & 3.57 \\
Unidepthv2
& \cellcolor{yellow!30}0.099 & \cellcolor{yellow!30}0.136
& \cellcolor{yellow!30}0.137 & 0.097
& \cellcolor{yellow!30}26.01 & \cellcolor{yellow!30}48.48
& \cellcolor{yellow!30}79.61 & \cellcolor{yellow!30}2.43
& 1.059 & 2.644 & 0.583 & 0.290
& \cellcolor{yellow!30}19.53 & \cellcolor{yellow!30}37.37
& 54.69 & 4.86 \\
Metric3D v2
& \cellcolor{green!30}0.073 & \cellcolor{green!30}0.106
& \cellcolor{green!30}0.104 & \cellcolor{green!30}0.068
& \cellcolor{green!30}31.00 & \cellcolor{green!30}56.00
& \cellcolor{green!30}90.42 & \cellcolor{green!30}1.00
& 1.126 & 1.854 & 0.513 & 0.434
& \cellcolor{green!30}20.57 & \cellcolor{green!30}41.99
& \cellcolor{yellow!30}65.95 & 4.00 \\
\bottomrule
\end{tabular}}
\end{table}

%% file: tables/main_mvs.tex
\begin{table}[t]
\centering
\caption{\textbf{Quantitative results for video-level geometry estimation.} pi3x and MapAnything also support depth-prior conditioning; their results are reported in Table~\ref{tab:mvd_completion}.}
\label{tab:multi_view}
\label{tab:multi_view_normal}
\label{tab:multi_view_challenge}
\compactspacing
\resizebox{\textwidth}{!}{%
\begin{tabular}{lcccccccc|cccccccc}
\toprule
& \multicolumn{8}{c|}{Normal}
& \multicolumn{8}{c}{Challenging} \\
\cmidrule(lr){2-9}\cmidrule(lr){10-17}
Model
& AE$\downarrow$ & RE$\downarrow$ & LE$\downarrow$ & RSD$\downarrow$
& $\delta_{1.05}\uparrow$ & $\delta_{1.10}\uparrow$
& $\delta_{1.25}\uparrow$ & Rank$\downarrow$
& AE$\downarrow$ & RE$\downarrow$ & LE$\downarrow$ & RSD$\downarrow$
& $\delta_{1.05}\uparrow$ & $\delta_{1.10}\uparrow$
& $\delta_{1.25}\uparrow$ & Rank$\downarrow$ \\
\midrule
MapAnything
& 0.295 & 0.404 & 0.312 & \cellcolor{blue!20}0.061
& 9.14 & 18.79 & 37.66 & 5.57
& 0.560 & 1.750 & 0.440 & \cellcolor{blue!20}0.190
& 4.90 & 8.40 & 43.40 & 4.43 \\
MASt3R
& 0.126 & 0.188 & 0.207 & 0.088
& 16.26 & 31.90 & 62.48 & 4.57
& 1.036 & 2.206 & 0.677 & 0.515
& 11.47 & 18.59 & 38.26 & 5.57 \\
pi3x
& 0.129 & 0.184 & 0.206 & \cellcolor{yellow!30}0.037
& 18.35 & 34.08 & 60.80 & 4.00
& \cellcolor{yellow!30}0.185 & \cellcolor{green!30}0.402
& \cellcolor{yellow!30}0.254 & \cellcolor{green!30}0.136
& \cellcolor{blue!20}11.48 & \cellcolor{blue!20}25.38
& \cellcolor{blue!20}63.09 & \cellcolor{yellow!30}2.14 \\
MUSt3R
& \cellcolor{blue!20}0.120 & \cellcolor{blue!20}0.174
& \cellcolor{blue!20}0.193 & 0.095
& \cellcolor{yellow!30}22.71 & \cellcolor{blue!20}38.96
& \cellcolor{blue!20}67.50 & \cellcolor{blue!20}3.29
& 0.996 & 2.116 & 0.671 & 0.497 & 10.69 & 19.36 & 41.99 & 4.86 \\
CUT3R
& \cellcolor{yellow!30}0.110 & \cellcolor{yellow!30}0.155
& \cellcolor{yellow!30}0.156 & 0.090
& \cellcolor{blue!20}22.65 & \cellcolor{yellow!30}42.39
& \cellcolor{yellow!30}76.12 & \cellcolor{yellow!30}2.57
& \cellcolor{blue!20}0.330 & \cellcolor{blue!20}0.646
& \cellcolor{blue!20}0.342 & 0.310
& \cellcolor{yellow!30}17.49 & \cellcolor{yellow!30}33.85
& \cellcolor{yellow!30}63.27 & \cellcolor{blue!20}2.71 \\
Depth Anything 3
& \cellcolor{green!30}0.078 & \cellcolor{green!30}0.108
& \cellcolor{green!30}0.102 & \cellcolor{green!30}0.029
& \cellcolor{green!30}30.52 & \cellcolor{green!30}54.16
& \cellcolor{green!30}92.25 & \cellcolor{green!30}1.00
& \cellcolor{green!30}0.167 & \cellcolor{yellow!30}0.432
& \cellcolor{green!30}0.219 & \cellcolor{yellow!30}0.174
& \cellcolor{green!30}21.16 & \cellcolor{green!30}38.66
& \cellcolor{green!30}81.56 & \cellcolor{green!30}1.29 \\
\bottomrule
\end{tabular}}
\end{table}

%% file: tables/main_mono_cmpl.tex
\begin{table}[t]
\centering
\caption{\textbf{Quantitative results for image-level geometry completion with sensor depth prior.} We input the sensor depth as the prior without any preprocessing.}
\label{tab:completion}
\label{tab:completion_normal}
\label{tab:completion_challenge}
\compactspacing
\resizebox{\textwidth}{!}{%
\begin{tabular}{lcccccccc|cccccccc}
\toprule
& \multicolumn{8}{c|}{Normal}
& \multicolumn{8}{c}{Challenging} \\
\cmidrule(lr){2-9}\cmidrule(lr){10-17}
Model
& AE$\downarrow$ & RE$\downarrow$ & LE$\downarrow$ & RSD$\downarrow$
& $\delta_{1.05}\uparrow$ & $\delta_{1.10}\uparrow$
& $\delta_{1.25}\uparrow$ & Rank$\downarrow$
& AE$\downarrow$ & RE$\downarrow$ & LE$\downarrow$ & RSD$\downarrow$
& $\delta_{1.05}\uparrow$ & $\delta_{1.10}\uparrow$
& $\delta_{1.25}\uparrow$ & Rank$\downarrow$ \\
\midrule
PromptDA
& 0.081 & 0.111 & 0.108 & 0.145 & 54.26 & 72.88 & 87.75 & 9.71
& \cellcolor{yellow!30}1.412 & \cellcolor{green!30}2.610
& \cellcolor{blue!20}0.871 & 0.727 & 9.90 & 17.84 & 28.91 & 5.86 \\
Marigold-dc
& 0.077 & 0.112 & 0.079 & 0.153 & 63.38 & 79.38 & 92.33 & 9.29
& 3.813 & 8.175 & 1.296 & 0.783 & 12.04 & 18.29 & 29.23 & 8.71 \\
CDM-L515
& 0.046 & 0.057 & 0.056 & \cellcolor{yellow!30}0.050
& 64.14 & 86.99 & 97.10 & 7.00
& \cellcolor{blue!20}2.569 & 7.541 & \cellcolor{yellow!30}0.759
& \cellcolor{blue!20}0.672 & \cellcolor{blue!20}26.43
& \cellcolor{yellow!30}38.93 & \cellcolor{yellow!30}50.59
& \cellcolor{yellow!30}3.00 \\
CDM-D435
& 0.043 & 0.053 & 0.051 & \cellcolor{blue!20}0.052
& 71.57 & 87.69 & 96.69 & 6.43
& 2.721 & \cellcolor{blue!20}5.079 & 0.907
& \cellcolor{yellow!30}0.462 & \cellcolor{yellow!30}26.69
& \cellcolor{blue!20}37.24 & \cellcolor{blue!20}48.37
& \cellcolor{blue!20}3.14 \\
DepthLab
& 0.031 & 0.046 & 0.044 & 0.064
& 69.65 & 89.14 & \cellcolor{blue!20}98.70 & 5.57
& 4.010 & 8.394 & 1.237 & 0.837 & 17.06 & 24.02 & 34.38 & 8.71 \\
PriorDA
& 0.037 & 0.048 & 0.041 & 0.067
& \cellcolor{blue!20}78.54 & \cellcolor{yellow!30}93.61
& 98.52 & 5.00
& 2.602 & 5.383 & 0.990 & 0.730
& 20.83 & 30.27 & 45.31 & 4.86 \\
InfiniDepth
& 0.028 & 0.041 & 0.039 & 0.057
& 77.55 & 91.84 & 98.33 & 4.86
& 3.129 & 6.066 & 1.072 & 0.705 & 26.37 & 33.40 & 43.55 & 4.86 \\
LDCM
& \cellcolor{yellow!30}0.025 & \cellcolor{blue!20}0.038
& \cellcolor{yellow!30}0.036 & 0.056
& \cellcolor{yellow!30}78.95 & \cellcolor{blue!20}92.82
& 98.66 & \cellcolor{blue!20}3.00
& 3.975 & 8.158 & 1.209 & 0.897
& 21.68 & 27.60 & 37.43 & 7.43 \\
Omni-dc
& \cellcolor{yellow!30}0.025 & \cellcolor{yellow!30}0.037
& \cellcolor{yellow!30}0.036 & 0.055
& 77.98 & 92.72
& \cellcolor{yellow!30}98.85 & \cellcolor{yellow!30}2.86
& 3.739 & 7.817 & 1.185 & 0.845 & 19.34 & 26.17 & 37.17 & 7.29 \\
Lingbot-Depth
& \cellcolor{green!30}0.022 & \cellcolor{green!30}0.029
& \cellcolor{green!30}0.028 & \cellcolor{green!30}0.034
& \cellcolor{green!30}84.52 & \cellcolor{green!30}98.46
& \cellcolor{green!30}99.81 & \cellcolor{green!30}1.00
& \cellcolor{green!30}1.045 & \cellcolor{yellow!30}2.628
& \cellcolor{green!30}0.576 & \cellcolor{green!30}0.386
& \cellcolor{green!30}47.92 & \cellcolor{green!30}58.98
& \cellcolor{green!30}67.12 & \cellcolor{green!30}1.14 \\
\bottomrule
\end{tabular}}
\end{table}

%% file: tables/main_mvs_cmpl.tex
\begin{table}[t]
\centering
\caption{\textbf{Quantitative results for video-level geometry completion with sensor depth prior.} We input the sensor depth for every view as the prior without any preprocessing. ``CAPA\_{*}'' means CAPA framework supported by ``*'' pretrained model.}
\label{tab:mvd_completion}
\label{tab:mvd_normal}
\label{tab:mvd_challenge}
\compactspacing
\resizebox{\textwidth}{!}{%
\begin{tabular}{lcccccccc|cccccccc}
\toprule
& \multicolumn{8}{c|}{Normal}
& \multicolumn{8}{c}{Challenging} \\
\cmidrule(lr){2-9}\cmidrule(lr){10-17}
Model
& AE$\downarrow$ & RE$\downarrow$ & LE$\downarrow$ & RSD$\downarrow$
& $\delta_{1.05}\uparrow$ & $\delta_{1.10}\uparrow$
& $\delta_{1.25}\uparrow$ & Rank$\downarrow$
& AE$\downarrow$ & RE$\downarrow$ & LE$\downarrow$ & RSD$\downarrow$
& $\delta_{1.05}\uparrow$ & $\delta_{1.10}\uparrow$
& $\delta_{1.25}\uparrow$ & Rank$\downarrow$ \\
\midrule
pi3x
& 0.097 & 0.131 & 0.139 & 0.062 & 26.43 & 44.76 & 80.46 & 5.00
& 2.811 & 5.781 & 1.079 & 0.911 & 8.42 & 15.91 & 37.83 & 4.43 \\
MapAnything
& 0.028 & 0.040 & 0.038 & 0.053 & 77.44 & 93.67 & 98.76 & 4.00
& 3.334 & 7.172 & 1.117 & 0.839
& 20.96 & \cellcolor{blue!20}28.61
& \cellcolor{blue!20}39.64 & 4.14 \\
CAPA-MoGe2
& \cellcolor{blue!20}0.019 & \cellcolor{yellow!30}0.027
& \cellcolor{yellow!30}0.027 & \cellcolor{green!30}0.027
& \cellcolor{blue!20}85.42 & \cellcolor{blue!20}96.95
& \cellcolor{yellow!30}99.76 & \cellcolor{blue!20}2.29
& \cellcolor{green!30}2.351 & \cellcolor{green!30}4.692
& \cellcolor{yellow!30}1.037 & \cellcolor{green!30}0.524
& \cellcolor{blue!20}21.43 & 26.26 & 37.07
& \cellcolor{yellow!30}2.43 \\
CAPA-Unidepthv2
& \cellcolor{yellow!30}0.018 & \cellcolor{yellow!30}0.027
& \cellcolor{yellow!30}0.027 & \cellcolor{green!30}0.027
& \cellcolor{yellow!30}86.15 & \cellcolor{yellow!30}97.25
& \cellcolor{yellow!30}99.76 & \cellcolor{yellow!30}1.86
& \cellcolor{blue!20}2.522 & \cellcolor{blue!20}4.99
& \cellcolor{blue!20}1.041 & \cellcolor{blue!20}0.607
& \cellcolor{yellow!30}25.69 & \cellcolor{yellow!30}31.94
& \cellcolor{yellow!30}41.97 & \cellcolor{blue!20}2.57 \\
CAPA-VGGT
& \cellcolor{green!30}0.016 & \cellcolor{green!30}0.022
& \cellcolor{green!30}0.022 & \cellcolor{blue!20}0.028
& \cellcolor{green!30}91.44 & \cellcolor{green!30}98.71
& \cellcolor{green!30}99.89 & \cellcolor{green!30}1.29
& \cellcolor{yellow!30}2.352 & \cellcolor{yellow!30}4.811
& \cellcolor{green!30}0.974 & \cellcolor{yellow!30}0.604
& \cellcolor{green!30}31.48 & \cellcolor{green!30}38.14
& \cellcolor{green!30}44.87 & \cellcolor{green!30}1.43 \\
\bottomrule
\end{tabular}}
\end{table}

%% file: sections/06_insights.tex
\section{Comprehensive Insights} \label{sec:compre_insights}

We first examine the coupling between predicted depth and camera intrinsics. We then analyze the performance characteristics of different 3D foundation models under our unified point-pair distance evaluation protocol. Unless otherwise specified, we use the normal subset to characterize general performance and both subsets for robustness analysis. Additional qualitative results are provided in the appendix ~\ref{app:qualitative_results}.

\begin{table}[t]
  \centering
  \caption{Average rank (Rank $\downarrow$) of each model on the depth benchmark
           (12 metrics) and the intrinsics benchmark (24 metrics).
           \textbf{Bold}: best; \underline{underlined}: second best.}
  \label{tab:rank_summary}
  \begin{tabular}{l c c c c c}
    \toprule
    Benchmark & Depth Pro & MoGe-2 & MetricAnything & Unidepth & Unidepthv2 \\
    \midrule
    Depth benchmark      & 4.25 & 2.83 & \underline{2.75} & \textbf{2.08} & 2.83 \\
    Intrinsics benchmark & 2.89 & \textbf{1.00} & \underline{1.50} & 3.46 & 3.83 \\
    \midrule
    Overall rank         & 3.57 & \textbf{1.92} & \underline{2.13} & 2.77 & 3.33 \\
    \bottomrule
  \end{tabular}
\end{table}

\textbf{Direct evaluation of point-pair distance estimation.} To investigate why direct evaluation is necessary, we conduct two experiments. First, we evaluate five image-level geometry estimation models on standard depth and camera-intrinsics benchmarks. Table~\ref{tab:rank_summary} summarizes their average ranks (see full results in appendix~\ref{app:detailed_coupling}). The rankings exhibit obvious task-specific discrepancies: Unidepth performs best on depth but relatively poorly on intrinsics, whereas MoGe-2 shows the opposite trend. Moreover, despite achieving the best combined depth--intrinsics rank, MoGe-2 is outperformed by Unidepthv2 on Pumpire (Table~\ref{tab:mono}). Second, under the Pumpire evaluation protocol, we replace each model's predicted intrinsics with ground-truth intrinsics when back-projecting its predicted depth, which degrades performance in most cases. These results demonstrate that point-pair distance estimation depends on depth--intrinsics coupling and cannot be reliably characterized by separate benchmarks, motivating its direct evaluation.

\begin{finding}\label{find:tasks_coupling}
    The coupling between predicted depth and camera intrinsics renders their separate evaluation insufficient for reliably characterizing point-pair distance estimation.
\end{finding}

\begin{figure}[t]
    \centering
    \includegraphics[width=\linewidth]{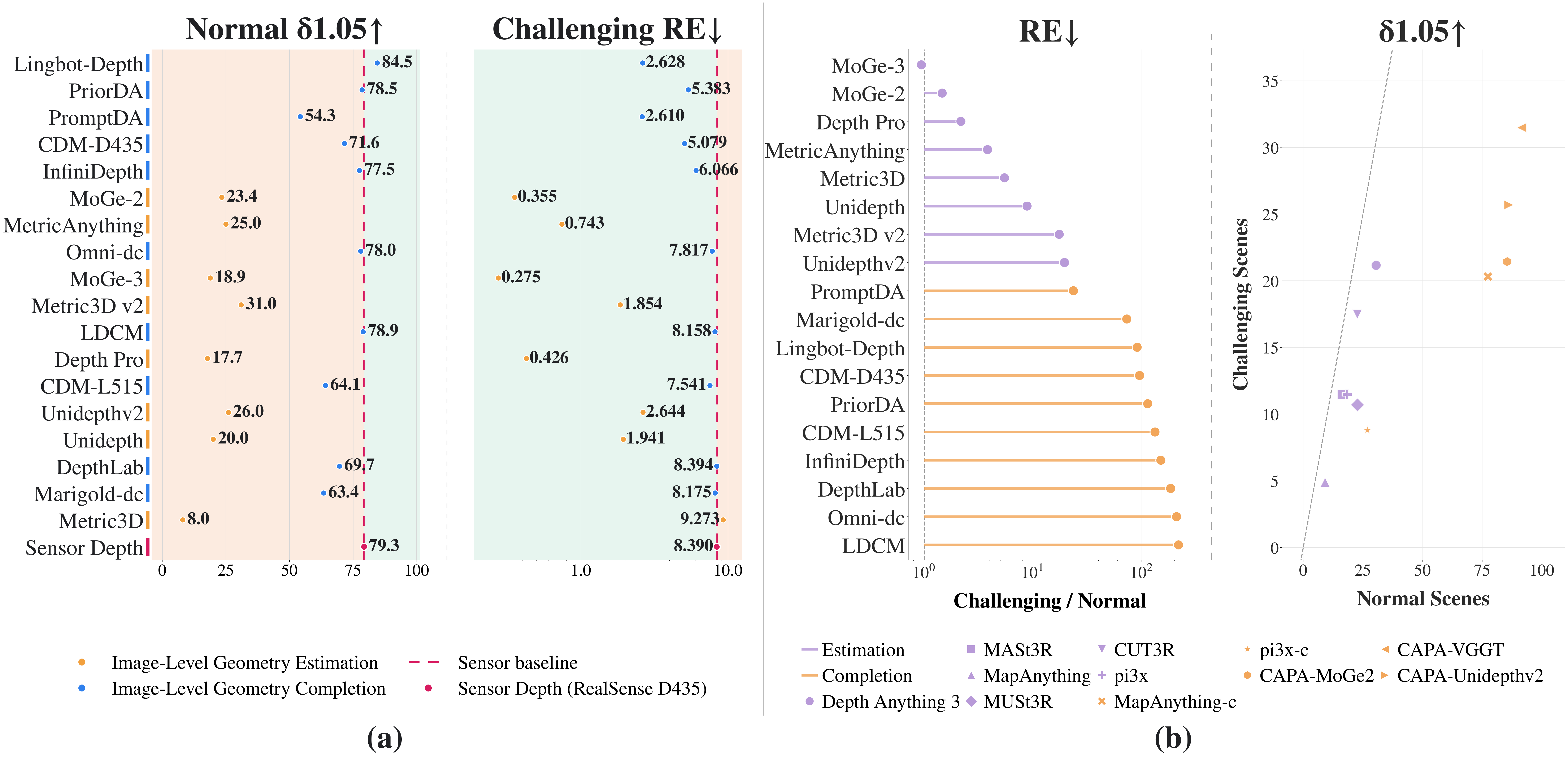}
    \caption{(a) Image-level geometry estimation (orange) and completion (blue) versus raw RealSense D435 depth: $\delta_{1.05}$ on normal scenes and RE on challenging scenes. Red dashed lines mark the sensor baseline; green regions indicate better performance. (b) Robustness of geometry estimation (purple) and completion (orange): challenging-to-normal RE ratios for image-level models (left), and $\delta_{1.05}$ on normal (x-axis) versus challenging (y-axis) scenes for multi-view models (right). Lower RE ratios indicate less degradation; points closer to the dashed $y=x$ line indicate more consistent accuracy across scene difficulty. The suffix “-c” in MapAnything-c and pi3x-c denotes their geometry completion settings with depth priors.}
    \label{fig:sensor_vs_models_and_robustness}
\end{figure}

\textbf{Sensor depth versus learned 3D foundation models.} We provide the first quantified comparison between raw sensor depth and learned 3D foundation models for point-pair distance estimation. On normal scenes (Figure~\subfigref{fig:sensor_vs_models_and_robustness}{a}), raw RealSense D435 depth is a remarkably strong baseline: only 5 of the 18 baseline configurations approach or surpass its $\delta_{1.05}$, all of them completion models (Lingbot-Depth, PriorDA, Omni-dc, LDCM, and InfiniDepth), while the best estimation model reaches barely half its accuracy. The ordering reverses on challenging scenes, where 16 of the 18 baseline configurations achieve a lower RE than the sensor: optical sensitivity and structural vulnerability invalidate active measurement, whereas learned models fall back on geometric priors.

\begin{finding} \label{find:sensor_vs_model}
    Raw sensor depth is highly reliable on benign surfaces, but under challenging sensing conditions learned 3D foundation models are far more robust and surpass it.
\end{finding}

\textbf{Distinct robustness patterns of estimation and completion.} Figure~\subfigref{fig:sensor_vs_models_and_robustness}{b} contrasts the two families across scene difficulty. The challenging-to-normal RE ratio (left) yields a non-overlapping separation: estimation models stay within $1$--$20\times$, whereas completion
models degrade by $20$--$210\times$; the multi-view scatter (right) shows the same contrast, with estimation models lying near the $y=x$ line and completion models far below it. Yet larger degradation does not imply inferiority, as the completion cluster still attains the highest $\delta_{1.05}$ on both subsets. The two families therefore differ in robustness patterns: completion models are precise on most frames but suffer catastrophic outliers on optically and structurally degraded regions that inflate RE, whereas estimation models trade peak accuracy for uniformly bounded errors.

\begin{finding} \label{find:esti_vs_comp}
    Completion and estimation models exhibit distinct robustness patterns: the former attains higher peak accuracy but degrades catastrophically on challenging scenes, the latter keeps errors bounded throughout.
\end{finding}

\begin{figure}[t]
    \centering
    \includegraphics[width=\linewidth]{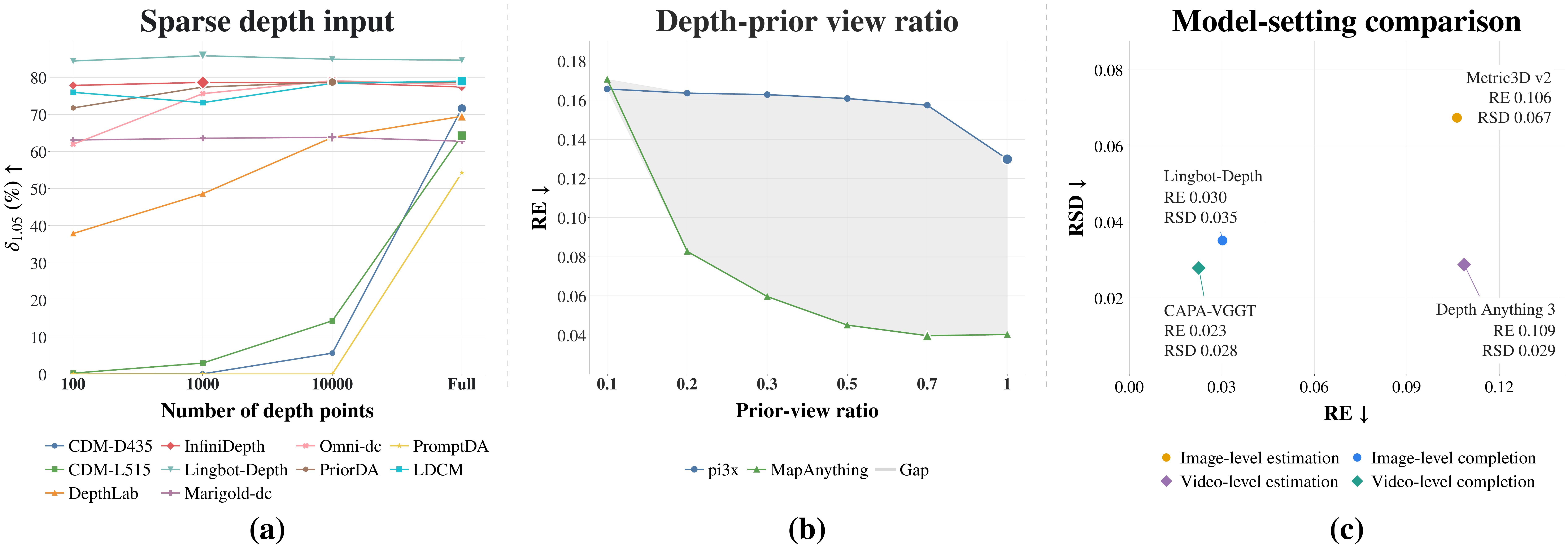}
    \caption{\textbf{(a)} Image-level completion accuracy ($\delta_{1.05}$) with 100, 1,000, or 10,000 randomly sampled depth points, or full sensor depth (Full).
  \textbf{(b)} Video-level completion RE as the fraction of views with depth priors increases from 0.1 to 1.0; shading indicates the gap between models. Enlarged markers in (a,b) highlight each model's best setting.
  \textbf{(c)} RE versus RSD of predicted distances for representative image- and video-level estimation and completion
  models. Points closer to the lower-left corner indicate greater accuracy and cross-frame consistency.}
    \label{fig:sparse_ratio_family}
\end{figure}

\textbf{Sparsely-sampled depth prior for image-level estimation models.} We randomly sample 100, 1{,}000, or 10{,}000 pixels from each sensor depth map as input priors, using the full depth map as a reference. Figure~\subfigref{fig:sparse_ratio_family}{a} reveals a counterintuitive result: increasing the number of depth points does not consistently improve completion accuracy, and several models perform best at intermediate densities. Completion models inherently face two distinct challenges: sparsity and sensor noise. Sparse inputs contain fewer geometric constraints but exclude much of the measurement noise, whereas full depth maps provide denser observations while retaining the complete sensor-noise pattern. Thus, performance is jointly determined by the model's robustness to both sparsity and noise.

\begin{finding}
      Geometry completion requires jointly addressing input sparsity and sensor noise; consequently, denser depth priors do not necessarily yield better performance.
\end{finding}

\textbf{Offering models with different numbers of depth priors.} We ablate the fraction of views with sensor-depth priors as $\rho\in\{0.1,0.2,0.3,0.5,0.7,1.0\}$. Since MapAnything~\cite{keetha2026mapanything} and pi3x~\cite{wang2026pi} are the only evaluated models that support partial per-view depth conditioning, we report results for these two methods. As shown in Figure~\subfigref{fig:sparse_ratio_family}{b}, RE decreases consistently for both models as $\rho$ increases, demonstrating the benefit of additional depth priors. The improvement is substantially larger for MapAnything, whereas pi3x exhibits a more gradual reduction. This contrast indicates that models differ considerably in their ability to integrate metric cues across views.

\begin{finding}
    Additional depth priors generally improve video-level completion, but the benefit depends on the model's ability to integrate metric information across views.
\end{finding}

\textbf{Comparison across model settings.} We analyze RE and RSD of the best-performing model from each of the four settings on the normal subset (Figure~\subfigref{fig:sparse_ratio_family}{c}). Moving from image-level estimation (Metric3D v2) to video-level estimation (Depth Anything 3) yields comparable RE (0.106 vs.\ 0.109) but substantially lower RSD (0.067 vs.\ 0.029), suggesting that multi-view models achieve better cross-view consistency. In contrast, the image-level completion model (Lingbot-Depth) achieves substantially lower RE than the estimation models, while attaining an RSD comparable to that of the video-level models (0.035 vs.\ 0.028 and 0.029, respectively), suggesting that per-view metric depth priors anchor independently predicted geometry to a consistent physical scale, thereby improving both metric accuracy and cross-view consistency. Video-level completion (CAPA-VGGT), which combines depth priors with multi-view context, achieves the lowest RE and RSD overall.

\begin{finding}
    Multi-view estimation improves consistency, depth-conditioned completion improves both accuracy and consistency, and combining both performs best overall.
\end{finding}

%% file: sections/07_conclusion.tex
\section{Conclusion}

We introduced Pumpire, a unified benchmark for directly evaluating the metric point-pair distance estimation capability of 3D foundation models. To support this evaluation, we constructed pumpire-6k with 100 real-world scenes and 6,400 frames annotated with physically measured distances, spanning diverse indoor and outdoor environments, surface properties, and viewing conditions. Our protocol enables unified comparison across image- and video-level geometry estimation and completion models. We systematically evaluate 29 baseline configurations and conduct in-depth analyses, revealing the necessity of direct distance evaluation, distinct accuracy--robustness trade-offs across model families and sensing conditions, and the effects of multi-view context and depth priors on distance accuracy and cross-view consistency.

%% file: sections/10_appendix.tex



\appendix
\section{Pixel-coordinate Annotation Pipeline}
  \label{app:annotation_pipeline}

\begin{figure}[t]
    \centering
    \includegraphics[width=\linewidth]{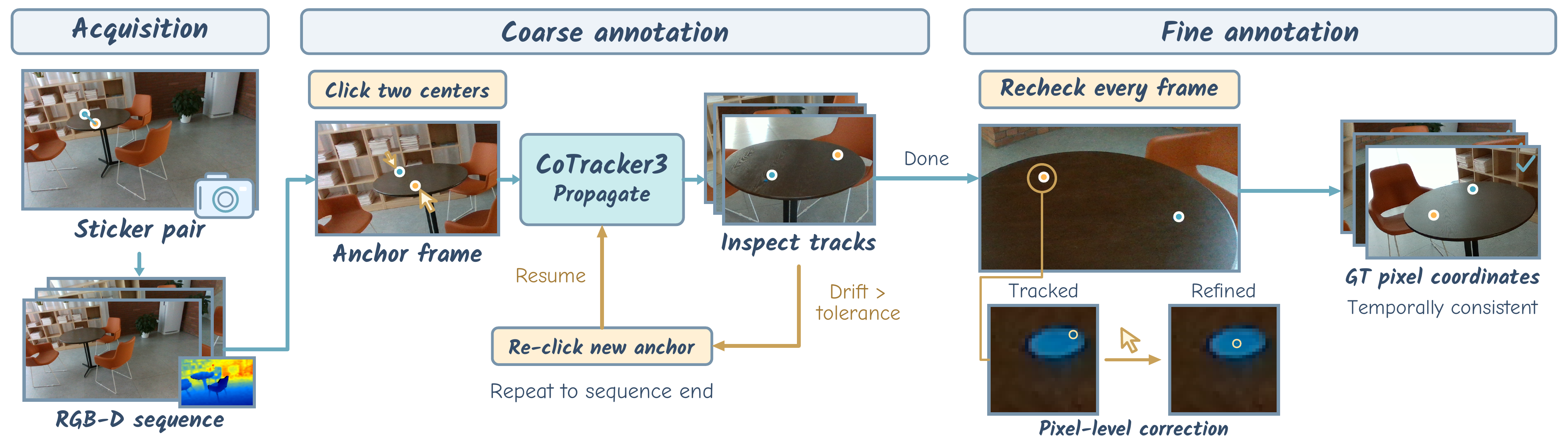}
    \caption{\textbf{Left}: Overview of our data acquisition pipeline, including scene preparation, multi-view RGB-D capture, and the generation of raw data for subsequent annotation. \textbf{Mid \& Right}: Overview of the coarse-to-fine data annotation pipeline. Valid frames are first sub-sampled from the raw image sequence, followed by anchor-frame annotation, automatic keypoint tracking, deviation-based re-annotation, and frame-by-frame verification to produce the final annotated sequence.}
    \label{fig:acquisition_annotation}
\end{figure}

We propose a lightweight coarse-to-fine annotation pipeline (Figure~\ref{fig:acquisition_annotation}) for per-frame pixel-coordinate annotation of sticker centers, substantially reducing manual annotation effort through a semi-automated human-in-the-loop framework built upon CoTracker3~\cite{karaev2025cotracker3}. In the coarse stage, the user manually clicks the centers of the two stickers on an anchor frame to initialize their pixel coordinates. These coordinates are provided to CoTracker3 as query points, which are then propagated to subsequent frames to obtain the corresponding pixel coordinates of the sticker centers throughout the sequence. The propagated trajectories are visually inspected, and whenever the tracking drift exceeds an acceptable tolerance, the current frame is selected as a new anchor, the sticker centers are re-clicked, and trajectory propagation resumes until the end of the sequence. In the fine stage, every frame is revisited, and all propagated keypoints are manually refined to pixel-level accuracy. Owing to the strong temporal coherence between adjacent frames, even subtle localization errors are readily identifiable during sequential inspection, enabling efficient local corrections and ultimately producing accurate and temporally consistent ground-truth annotations.

\section{\texorpdfstring{Baseline Configurations}{Baseline Configurations}} \label{app:baselines}
As summarized in Table~\ref{tab:baselines}, we benchmark 26 representative method entries across four settings: 8 image-level estimation methods, 9 image-level completion methods, 6 video-level estimation methods, and 3 video-level completion methods. We exclusively consider metric 3D foundation models that recover scene geometry at absolute scale. The table groups variants from the same method family into a single entry; counting the two sensor-specific CDM variants and the three CAPA backbone variants separately yields the 29 baseline configurations evaluated in Section~\ref{sec:systematic_evaluation}.

\section{Video-level model sampling strategy.} Throughout our systematic evaluation, all video-level models use the same sampling procedure with a fixed random seed. Consequently, all models obtain the same sequence of multi-view samples, including identical frame groups, sample ordering, and within-group frame ordering. This deterministic sampling setup eliminates variation from random view selection and enables direct comparison across video-level models.

\section{Additional Details for Comparison across Model Settings}
To fairly compare image- and video-level models in Figure~\subfigref{fig:sparse_ratio_family}{c}, we evaluate Depth Anything 3 and CAPA-VGGT using the same multi-view samples and use their shared sampling records to determine the evaluation weight of each image. Accordingly, each Metric3D v2 and Lingbot-Depth prediction is counted as many times as its frame appears in the multi-view samples. This procedure requires no additional image-level inference and gives all four models the same 24,320 frame occurrences over the 76 normal scenes. We then compute RE and RSD from the aligned predictions within each scene, following Section~\ref{sec:benchmark_protocol}, and average the resulting scene-level metrics equally across scenes.

\begin{table}[h]
\centering
\caption{\textbf{Overview of the evaluated baselines.} The evaluated CDM variants and CAPA backbones are listed in parentheses.}
\label{tab:baselines}
\small
\setlength{\tabcolsep}{4pt}
\renewcommand{\arraystretch}{1.2}
\begin{tabular}{
>{\raggedright\arraybackslash}m{2.0cm}
>{\raggedright\arraybackslash}m{1.8cm}
>{\raggedright\arraybackslash}m{9cm}
}
\toprule
\textbf{Input} & \textbf{Task} & \textbf{Models} \\
\midrule

\multirow{2}{*}{\raisebox{-2.2\baselineskip}{\textbf{Image-level}}}
& Estimation &
Metric3D~\cite{yin2023metric3d},
Unidepth~\cite{piccinelli2024unidepth},
MoGe-2~\cite{wang2507moge},
MoGe-3~\cite{kong2026moge},
Depth Pro~\cite{bochkovskiy2025depth},
MetricAnything~\cite{ma2026metricanything},
Metric3D v2~\cite{hu2024metric3d},
Unidepthv2~\cite{piccinelli2025unidepthv2} \\

\cmidrule(l){2-3}

& Completion &
CDM~\cite{liu2025manipulation} (CDM-L515, CDM-D435),
DepthLab~\cite{liu2024depthlab},
PriorDA~\cite{wang2025depth},
Lingbot-Depth~\cite{tan2026masked},
InfiniDepth~\cite{yu2026infinidepth},
Omni-dc~\cite{zuo2025omni},
LDCM~\cite{yu2026large},
Marigold-dc~\cite{viola2025marigold},
PromptDA~\cite{lin2025prompting} \\

\midrule

\multirow{2}{*}{\raisebox{-2.1\baselineskip}{\textbf{Video-level}}}
& Estimation &
MapAnything~\cite{keetha2026mapanything},
MASt3R~\cite{leroy2024grounding},
MUSt3R~\cite{cabon2025must3r},
Depth Anything 3~\cite{lin2511depth},
pi3x~\cite{wang2026pi},
CUT3R~\cite{wang2025continuous} \\

\cmidrule(l){2-3}

& Completion &
MapAnything~\cite{keetha2026mapanything},
pi3x~\cite{wang2026pi},
CAPA~\cite{ke2026depth}
(CAPA-VGGT~\cite{wang2025vggt},
CAPA-Unidepthv2~\cite{piccinelli2025unidepthv2},
CAPA-MoGe2~\cite{wang2507moge}) \\

\bottomrule
\end{tabular}
\end{table}

\section{\texorpdfstring{Detailed Exploration Results for Direct Distance Evaluation}{Detailed Results for Direct Distance Evaluation}} \label{app:detailed_coupling}

\input{tables/insight1_gtfov}

The ability to measure point-pair distance depends jointly on accurate depth maps and reliable camera intrinsics. This raises a fundamental question: do models with strong performance on both depth estimation and camera intrinsics estimation also achieve superior point-pair distance estimation?

To answer this, we evaluate five image-level geometry estimation models~\cite{bochkovskiy2025depth, wang2507moge, ma2026metricanything, piccinelli2024unidepth, piccinelli2025unidepthv2} on standard benchmarks for both tasks. For depth, we report AbsRel, L1, and $\delta_{1.25}$ on NYUD~\cite{silberman2012indoor}, KITTI~\cite{geiger2013vision}, ETH3D~\cite{schops2017multi}, and DIODE~\cite{vasiljevic2019diode}; for intrinsics, we report $AUC@1/5/10^\circ$ on LaMAR~\cite{sarlin2022lamar}, MegaDepth~\cite{li2018megadepth}, Stanford2D3D~\cite{armeni2017joint}, and TartanAir~\cite{wang2020tartanair}, following GeoCalib~\cite{veicht2024geocalib}. Table~\ref{tab:rank_summary} summarizes the average rankings. Table~\ref{tab:dpt_bench} and Table~\ref{tab:fov_bench} provide the corresponding per-dataset results. The rankings reveal clear task-specific disparities: for example, Unidepth achieves the best depth rank but only the fourth-best intrinsics rank, whereas MoGe-2 exhibits the opposite trend, indicating that superiority in one task does not generalize to the other.

More importantly, even jointly considering the two benchmarks fails to predict point-pair distance estimation performance. Despite achieving the best overall depth--intrinsics ranking, MoGe-2 is consistently outperformed by Unidepthv2 on our distance estimation benchmark (Table~\ref{tab:mono}). To further investigate this discrepancy, we replace the predicted intrinsics with ground-truth intrinsics when back-projecting the predicted depth maps. To reveal general patterns, we conduct experiments on the normal subset only. Surprisingly, this substitution degrades across the majority of metrics --- 47 of the 56 model--metric pairs (Table~\ref{tab:mono_gtfov_normal}), demonstrating that predicted depth maps are intrinsically coupled with their corresponding predicted intrinsics. These results show that neither existing depth and intrinsics benchmarks, nor their simple combination, can faithfully characterize a model's point-pair distance estimation capability, highlighting the necessity of direct evaluation.

\begin{table*}[t]
  \centering
  \caption{\textbf{Depth benchmark results for image-level metric geometry estimation models.} The best values are highlighted in \colorbox{green!30}{green}, and the second-best ones in \colorbox{yellow!30}{yellow}.}
  \label{tab:dpt_bench}
  \scriptsize
  \setlength{\tabcolsep}{2pt}
  \begin{tabular}{l
    ccc|ccc|ccc|ccc}
    \toprule
    Model &
    \multicolumn{3}{c}{NYU-D} &
    \multicolumn{3}{c}{KITTI} &
    \multicolumn{3}{c}{DIODE} &
    \multicolumn{3}{c}{ETH3D} \\
    \cmidrule(lr){2-4} \cmidrule(lr){5-7} \cmidrule(lr){8-10} \cmidrule(lr){11-13}
    & AbsRel $\downarrow$ & L1 (m) $\downarrow$ & $\delta_{1.25} \uparrow$
    & AbsRel $\downarrow$ & L1 (m) $\downarrow$ & $\delta_{1.25} \uparrow$
    & AbsRel $\downarrow$ & L1 (m) $\downarrow$ & $\delta_{1.25} \uparrow$
    & AbsRel $\downarrow$ & L1 (m) $\downarrow$ & $\delta_{1.25} \uparrow$ \\
    \midrule
    Depth Pro
    & 0.09 & 0.25 & 0.93
    & 0.14 & 2.34 & 0.83
    & 0.40 & 4.46 & 0.41
    & 0.38 & 3.25 & 0.33 \\
    MoGe-2
    & 0.08 & 0.21 & 0.96
    & 0.21 & 3.59 & 0.45
    & \cellcolor{yellow!30}0.33 & \cellcolor{yellow!30}2.62 & 0.54
    & \cellcolor{green!30}0.10 & \cellcolor{green!30}0.62 & \cellcolor{yellow!30}0.88 \\
    MetricAnything
    & 0.10 & 0.27 & 0.94
    & \cellcolor{yellow!30}0.09 & 1.60 & 0.94
    & 0.34 & \cellcolor{green!30}2.49 & \cellcolor{yellow!30}0.65
    & \cellcolor{yellow!30}0.11 & \cellcolor{yellow!30}0.67 & \cellcolor{green!30}0.90 \\
    Unidepth
    & \cellcolor{green!30}0.06 & \cellcolor{green!30}0.14 & \cellcolor{green!30}0.98
    & \cellcolor{green!30}0.05 & \cellcolor{green!30}1.04 & \cellcolor{green!30}0.98
    & \cellcolor{green!30}0.27 & 2.64 & \cellcolor{green!30}0.67
    & 0.58 & 3.20 & 0.14 \\
    Unidepthv2
    & \cellcolor{yellow!30}0.07 & \cellcolor{yellow!30}0.18 & \cellcolor{yellow!30}0.96
    & 0.09 & \cellcolor{yellow!30}1.58 & \cellcolor{yellow!30}0.95
    & 0.78 & 7.07 & 0.54
    & 0.21 & 1.23 & 0.68 \\
    \bottomrule
  \end{tabular}
\end{table*}

\begin{table*}[t]
  \centering
  \caption{\textbf{Intrinsics benchmark results for image-level metric geometry estimation models.} $@1/5/10^\circ$ refer to $AUC@1/5/10^\circ$. vFoV, hFoV refer to vertical FoV and horizontal FoV. The best values are highlighted in \colorbox{green!30}{green}, and the second-best ones in \colorbox{yellow!30}{yellow}. Methods trained on evaluated datasets are in \textcolor{gray}{gray} and excluded from the ranking to ensure a fair comparison}
  \label{tab:fov_bench}
  \scriptsize
  \setlength{\tabcolsep}{2pt}

  \begin{subtable}[b]{0.48\textwidth}
    \centering
    \caption{LaMAR-2K}
    \begin{tabular}{l ccc ccc}
      \toprule
      Model & \multicolumn{3}{c}{vFoV} & \multicolumn{3}{c}{hFoV} \\
      \cmidrule(lr){2-4} \cmidrule(lr){5-7}
      & @$1^\circ$ & @$5^\circ$ & @$10^\circ$ & @$1^\circ$ & @$5^\circ$ & @$10^\circ$ \\
      \midrule
      Depth Pro
        & 0.121 & 0.235 & 0.377 & 0.136 & 0.260 & 0.447 \\
      MoGe-2
        & \cellcolor{green!30}0.243 & \cellcolor{green!30}0.526 & \cellcolor{green!30}0.727 & \cellcolor{green!30}0.287 & \cellcolor{green!30}0.581 & \cellcolor{green!30}0.766 \\
      MetricAnything
        & \cellcolor{yellow!30}0.224 & \cellcolor{yellow!30}0.493 & \cellcolor{yellow!30}0.698 & \cellcolor{yellow!30}0.262 & \cellcolor{yellow!30}0.548 & \cellcolor{yellow!30}0.743 \\
      Unidepth
        & 0.051 & 0.371 & 0.491 & 0.015 & 0.324 & 0.479 \\
      Unidepthv2
        & 0.040 & 0.123 & 0.312 & 0.055 & 0.193 & 0.423 \\
      \bottomrule
    \end{tabular}
  \end{subtable}
  \hfill
  \begin{subtable}[b]{0.48\textwidth}
    \centering
    \caption{MegaDepth-2K}
    \begin{tabular}{l ccc ccc}
      \toprule
      Model & \multicolumn{3}{c}{vFoV} & \multicolumn{3}{c}{hFoV} \\
      \cmidrule(lr){2-4} \cmidrule(lr){5-7}
      & @$1^\circ$ & @$5^\circ$ & @$10^\circ$ & @$1^\circ$ & @$5^\circ$ & @$10^\circ$ \\
      \midrule
      Depth Pro
        & \cellcolor{yellow!30}0.196 & \cellcolor{yellow!30}0.449 & \cellcolor{yellow!30}0.660 & \cellcolor{yellow!30}0.176 & \cellcolor{yellow!30}0.423 & \cellcolor{yellow!30}0.636 \\
      MoGe-2
        & \textcolor{gray}{0.176} & \textcolor{gray}{0.412} & \textcolor{gray}{0.618} & \textcolor{gray}{0.174} & \textcolor{gray}{0.397} & \textcolor{gray}{0.593} \\
      MetricAnything
        & \cellcolor{green!30}0.270 & \cellcolor{green!30}0.537 & \cellcolor{green!30}0.722 & \cellcolor{green!30}0.252 & \cellcolor{green!30}0.514 & \cellcolor{green!30}0.702 \\
      Unidepth
        & 0.068 & 0.162 & 0.274 & 0.058 & 0.149 & 0.253 \\
      Unidepthv2
        & \textcolor{gray}{0.124} & \textcolor{gray}{0.305} & \textcolor{gray}{0.494} & \textcolor{gray}{0.120} & \textcolor{gray}{0.289} & \textcolor{gray}{0.475} \\
      \bottomrule
    \end{tabular}
  \end{subtable}

  \vspace{1ex}
  \begin{subtable}[b]{0.48\textwidth}
    \centering
    \caption{Stanford2D3D}
    \begin{tabular}{l ccc ccc}
      \toprule
      Model & \multicolumn{3}{c}{vFoV} & \multicolumn{3}{c}{hFoV} \\
      \cmidrule(lr){2-4} \cmidrule(lr){5-7}
      & @$1^\circ$ & @$5^\circ$ & @$10^\circ$ & @$1^\circ$ & @$5^\circ$ & @$10^\circ$ \\
      \midrule
      Depth Pro
        & 0.058 & 0.157 & 0.263 & 0.050 & 0.134 & 0.233 \\
      MoGe-2
        & \cellcolor{green!30}0.094 & \cellcolor{green!30}0.269 & \cellcolor{green!30}0.487 & \cellcolor{green!30}0.087 & \cellcolor{green!30}0.249 & \cellcolor{green!30}0.454 \\
      MetricAnything
        & \cellcolor{yellow!30}0.078 & \cellcolor{yellow!30}0.194 & \cellcolor{yellow!30}0.326 & \cellcolor{yellow!30}0.062 & \cellcolor{yellow!30}0.164 & \cellcolor{yellow!30}0.289 \\
      Unidepth
        & 0.032 & 0.085 & 0.172 & 0.028 & 0.077 & 0.150 \\
      Unidepthv2
        & 0.048 & 0.132 & 0.247 & 0.043 & 0.116 & 0.224 \\
      \bottomrule
    \end{tabular}
  \end{subtable}
  \hfill
  \begin{subtable}[b]{0.48\textwidth}
    \centering
    \caption{TartanAir}
    \begin{tabular}{l ccc ccc}
      \toprule
      Model & \multicolumn{3}{c}{vFoV} & \multicolumn{3}{c}{hFoV} \\
      \cmidrule(lr){2-4} \cmidrule(lr){5-7}
      & @$1^\circ$ & @$5^\circ$ & @$10^\circ$ & @$1^\circ$ & @$5^\circ$ & @$10^\circ$ \\
      \midrule
      Depth Pro
        & \textcolor{gray}{0.556} & \textcolor{gray}{0.740} & \textcolor{gray}{0.859} & \textcolor{gray}{0.549} & \textcolor{gray}{0.730} & \textcolor{gray}{0.851} \\
      MoGe-2
        & \textcolor{gray}{0.009} & \textcolor{gray}{0.080} & \textcolor{gray}{0.314} & \textcolor{gray}{0.009} & \textcolor{gray}{0.074} & \textcolor{gray}{0.303} \\
      MetricAnything
        & \cellcolor{green!30}0.364 & \cellcolor{green!30}0.702 & \cellcolor{green!30}0.847 & \cellcolor{green!30}0.353 & \cellcolor{green!30}0.688 & \cellcolor{green!30}0.839 \\
      Unidepth
        & \cellcolor{yellow!30}0.008 & \cellcolor{yellow!30}0.016 & \cellcolor{yellow!30}0.030 & \cellcolor{yellow!30}0.006 & \cellcolor{yellow!30}0.015 & \cellcolor{yellow!30}0.031 \\
      Unidepthv2
        & 0.000 & 0.004 & 0.093 & 0.000 & 0.004 & 0.083 \\
      \bottomrule
    \end{tabular}
  \end{subtable}
\end{table*}

\section{\texorpdfstring{Effect of Multi-view Sampling Stride}{Effect of Multi-view Sampling Stride}}
\begin{figure}[h]
    \centering
    \includegraphics[width=\linewidth]{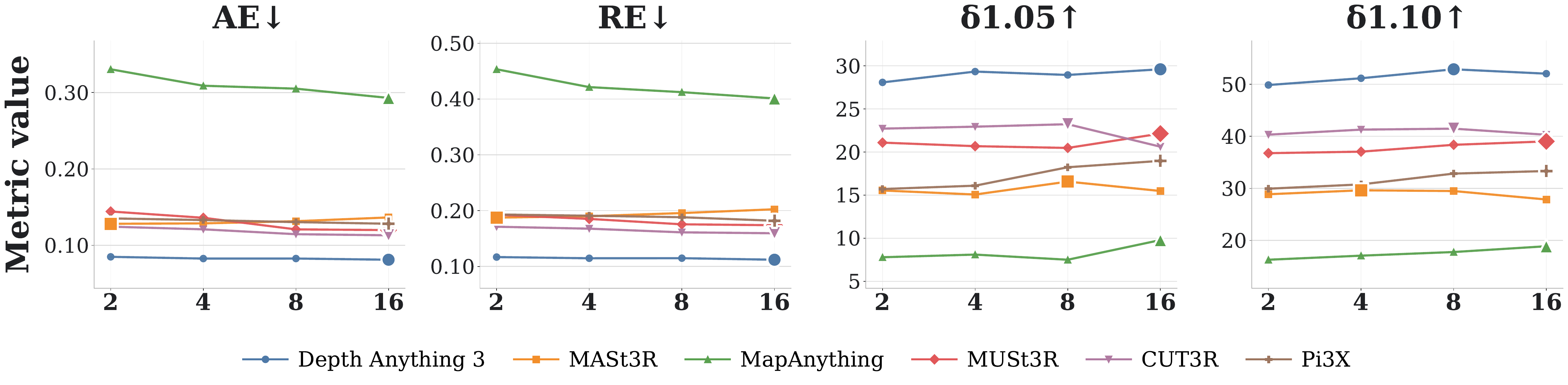}
    \caption{Effect of sampling stride on video-level geometry estimation models (normal subset
only). Each curve corresponds to one model, and the highlighted marker on each curve indicates
that model’s best-performing stride.}
    \label{fig:multiview_stride}
\end{figure}

To complement the video-level evaluation in Section~\ref{sec:compre_insights} and isolate the effect of inter-view
  baseline, we sample $N=4$ frames from each 64-frame sequence using a uniform
    stride $s\in\{2,4,8,16\}$. We draw 16 samples per scene and shuffle the sample to mitigate ordering bias. As shown in
    Figure~\ref{fig:multiview_stride}, wider sampling generally improves $\delta_{1.05}$: Depth Anything 3, MapAnything,
    MUSt3R, and pi3x achieve their best performance at $s=16$. MASt3R and CUT3R instead peak at $s=8$, indicating that the
    benefit of wider baselines can be offset by reduced visual overlap. Overall, sampling stride reflects a trade-off between geometric diversity and cross-view correspondence.

\begin{finding}
    Wider sampling strides generally benefit one-forward video-level geometry estimation models by providing larger inter-view baselines, while excessively wide spacing can reduce visual overlap and offset this gain.
\end{finding}

\section{\texorpdfstring{Additional Qualitative Results}{Additional Qualitative Results}} \label{app:qualitative_results}

We provide qualitative examples for three analyses in Section~\ref{sec:compre_insights}: depth--intrinsics coupling, sparsely sampled depth priors for image-level completion, and different fractions of depth-prior views for video-level completion. These examples complement the aggregate results in Figure~\ref{fig:sparse_ratio_family} and the direct-evaluation analysis in Finding~\ref{find:tasks_coupling}.


\begin{figure*}[h]
    \centering
    \includegraphics[width=\linewidth]{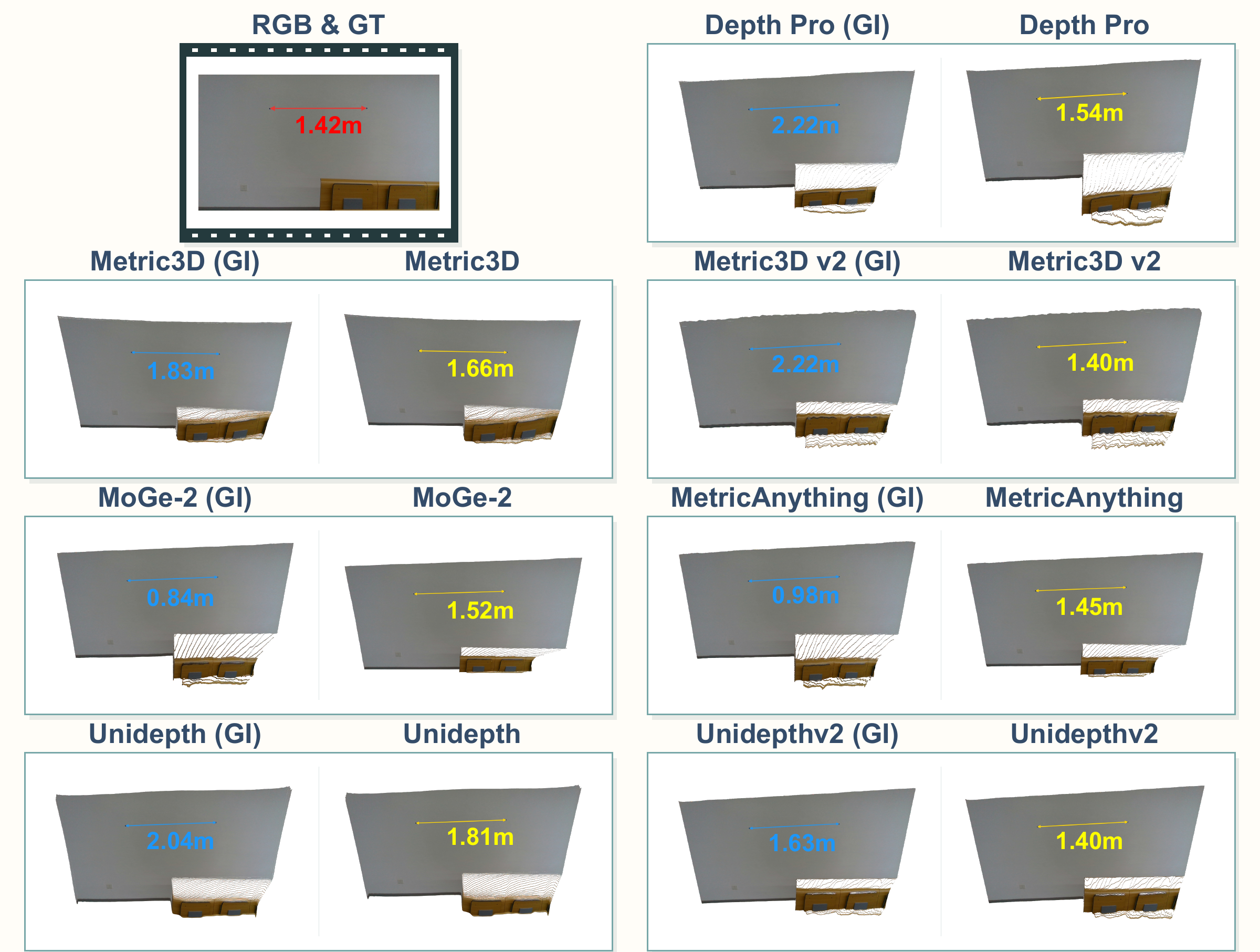}
    \caption{\textbf{Qualitative comparison for image-level geometry estimation models between back-projecting with ground-truth intrinsics and models' predicted intrinsics.} Ground-truth annotations are shown in \textcolor{red}{red}, (ground-truth intrinsics + depth maps) are shown in \textcolor{depthblue}{blue}, (predicted intrinsics + depth maps) are shown in \textcolor{yellow}{yellow}.}
    \label{fig:vis_gt_pred_fov}
\end{figure*}

\begin{figure*}[h]
    \centering
    \includegraphics[width=\linewidth]{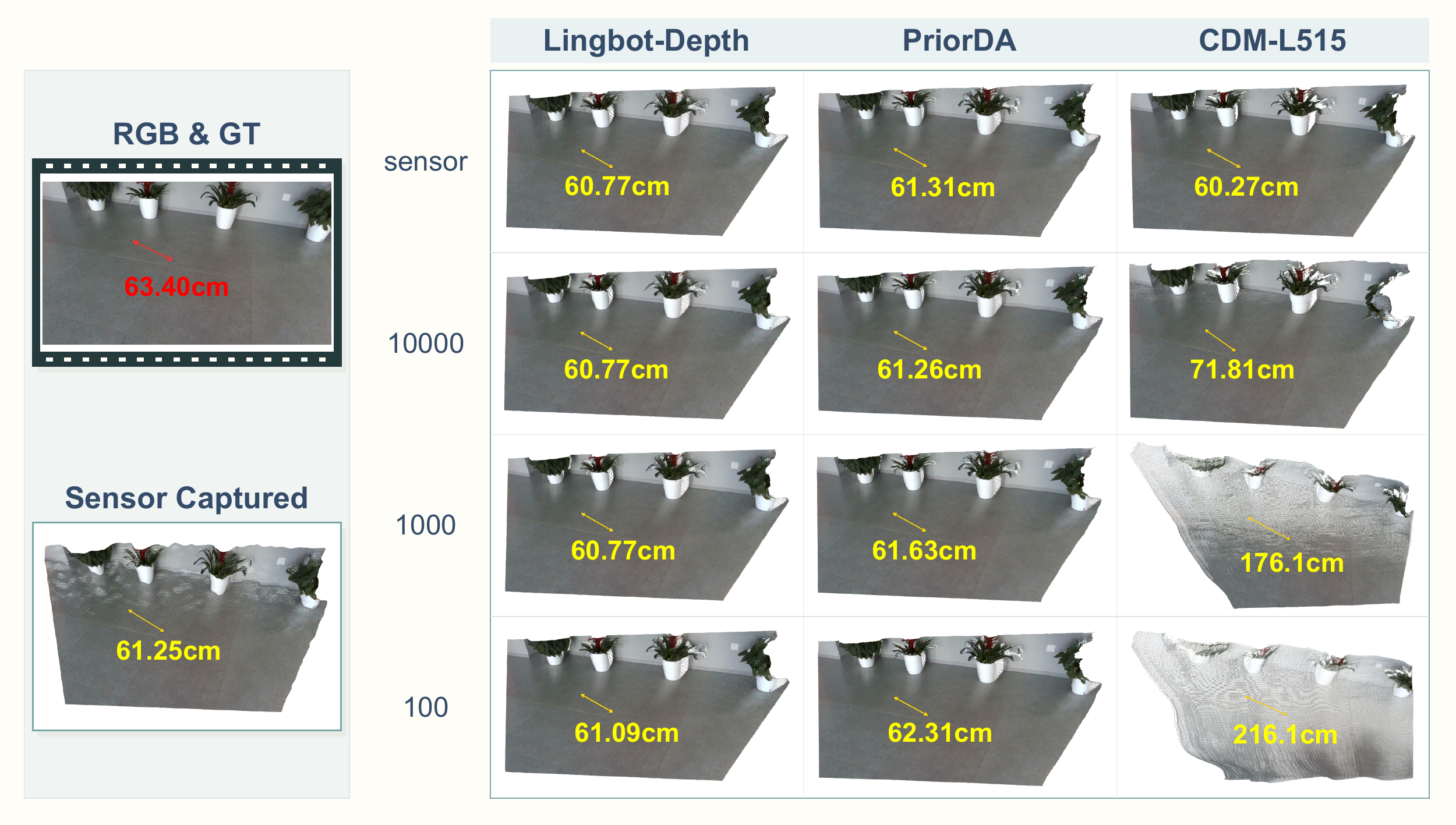}
    \caption{\textbf{Qualitative comparison for image-level geometry completion models on different prior depth patterns with varying sparsity levels.} The results showcase that Lingbot-Depth and PriorDA perform robustly under different patterns, while CDM-L515 exhibits a substantial performance drop due to the large pattern domain gap between its training data and the evaluated prior patterns.}
    \label{fig:sensor_vs_sparse}
\end{figure*}

\begin{figure}[h]
    \centering
    \includegraphics[width=\linewidth]{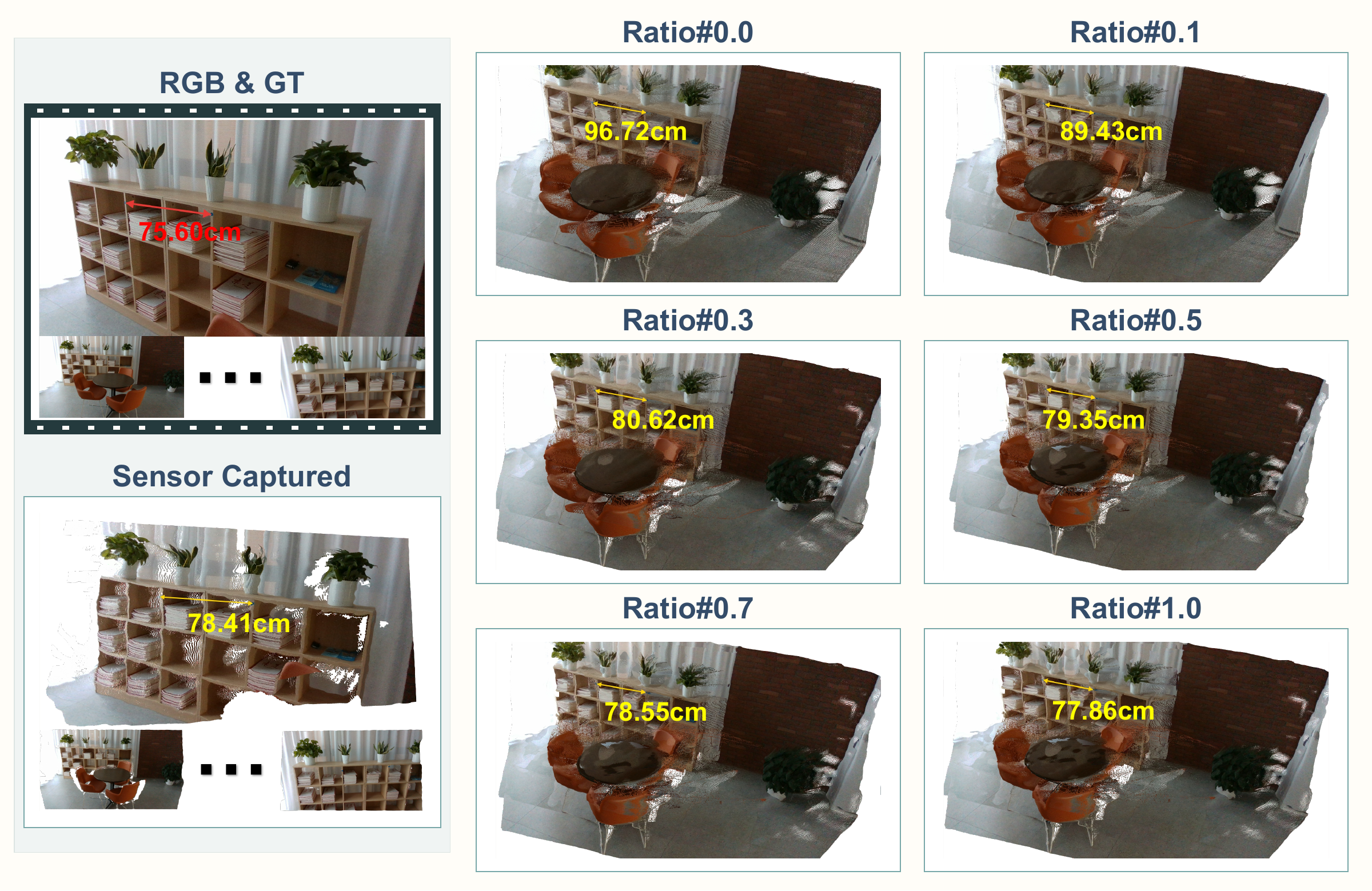}
    \caption{\textbf{Qualitative results for MapAnything~\cite{keetha2026mapanything} with varying fractions of views provided with depth priors.} The distance of ``Sensor Captured'' is obtained by averaging the sensor measured distances of the ten views sampled. The results demonstrate that, as the ratio of depth priors increased, MapAnything gains continuous performance improvement. ``Ratio\#x'' indicates the ratio of prior input to the model.}
    \label{fig:multiview_vis}
\end{figure}

%% file: tables/insight1_gtfov.tex


\begin{table}[t]
\centering
\caption{\textbf{Image-level geometry estimation using ground-truth intrinsics.} For the seven evaluation metrics, values in parentheses denote changes relative to the corresponding normal-subset results in Table~\ref{tab:mono}; \textcolor{green!60!black}{green} and \textcolor{red}{red} indicate improvements and degradations, respectively.}
\label{tab:mono_gtfov_normal}
\scriptsize
\setlength{\tabcolsep}{1.5pt}
\resizebox{\textwidth}{!}{%
\begin{tabular}{lcccccccc}
\toprule
Model & AE $\downarrow$ & RE$\downarrow$ & LE$\downarrow$ & RSD$\downarrow$ & ${\delta}_{1.05}\uparrow$ & ${\delta}_{1.10}\uparrow$ & ${\delta}_{1.25}\uparrow$ & Rank$\downarrow$ \\
\midrule
Metric3D
& 1.319 (\textcolor{red}{+0.139})
& 1.895 (\textcolor{red}{+0.200})
& 0.893 (\textcolor{red}{+0.065})
& 0.340 (\textcolor{green!60!black}{-0.007})
& 5.16 (\textcolor{red}{-2.86})
& 9.89 (\textcolor{red}{-4.79})
& 23.42 (\textcolor{red}{-4.40})
& 8.00 \\
Depth Pro
& 0.144 (\textcolor{red}{+0.009})
& 0.210 (\textcolor{red}{+0.014})
& 0.240 (\textcolor{red}{+0.006})
& 0.175 (\textcolor{red}{+0.021})
& 16.53 (\textcolor{red}{-1.21})
& 31.68 (\textcolor{red}{-1.30})
& 62.77 (\textcolor{red}{-2.53})
& 5.14 \\
Unidepth
& 0.205 (\textcolor{red}{+0.046})
& 0.287 (\textcolor{red}{+0.067})
& 0.246 (\textcolor{red}{+0.039})
& 0.121 (\textcolor{green!60!black}{-0.013})
& 14.68 (\textcolor{red}{-5.32})
& 27.22 (\textcolor{red}{-11.16})
& 52.55 (\textcolor{red}{-15.89})
& 6.86 \\
MoGe-2
& 0.172 (\textcolor{green!60!black}{-0.004})
& 0.229 (\textcolor{green!60!black}{-0.013})
& 0.208 (\textcolor{red}{+0.004})
& 0.095 (\textcolor{red}{+0.009})
& 17.00 (\textcolor{red}{-6.40})
& 31.50 (\textcolor{red}{-9.43})
& 62.29 (\textcolor{red}{-0.79})
& 4.86 \\
MoGe-3
& 0.185 (\textcolor{green!60!black}{-0.023})
& 0.253 (\textcolor{green!60!black}{-0.039})
& 0.220 (\textcolor{green!60!black}{-0.016})
& 0.093 (\textcolor{red}{+0.009})
& 17.19 (\textcolor{red}{-1.70})
& 33.68 (\textcolor{red}{-1.81})
& 63.34 (\textcolor{green!60!black}{+3.45})
& 4.43 \\
MetricAnything
& 0.150 (\textcolor{red}{+0.005})
& 0.195 (\textcolor{red}{+0.001})
& 0.183 (\textcolor{red}{+0.007})
& 0.098 (\textcolor{red}{+0.009})
& 20.19 (\textcolor{red}{-4.81})
& 37.34 (\textcolor{red}{-3.94})
& 66.86 (\textcolor{red}{-0.10})
& 3.29 \\
Unidepthv2
& 0.121 (\textcolor{red}{+0.022})
& 0.165 (\textcolor{red}{+0.029})
& 0.153 (\textcolor{red}{+0.016})
& 0.098 (\textcolor{red}{+0.001})
& 23.58 (\textcolor{red}{-2.43})
& 43.01 (\textcolor{red}{-5.47})
& 75.41 (\textcolor{red}{-4.20})
& 2.29 \\
Metric3D v2
& 0.095 (\textcolor{red}{+0.022})
& 0.138 (\textcolor{red}{+0.032})
& 0.126 (\textcolor{red}{+0.022})
& 0.067 (\textcolor{green!60!black}{-0.001})
& 26.09 (\textcolor{red}{-4.91})
& 47.94 (\textcolor{red}{-8.06})
& 83.98 (\textcolor{red}{-6.44})
& 1.00 \\
\bottomrule
\end{tabular}}
\end{table}
